\documentclass{article}
\usepackage{spconf,amsmath,graphicx}
\usepackage{amstext,amsthm,amssymb,curves,xcolor}
\usepackage[retainorgcmds]{IEEEtrantools}
\usepackage{algorithm}
\usepackage{algpseudocode}
\usepackage{subcaption}
\usepackage{array}
\usepackage{arydshln}
\usepackage{multicol}
\usepackage{multirow}
\graphicspath{{./compare_png/}}

\theoremstyle{plain}
\newtheorem{thm}{Theorem}[section]
\theoremstyle{definition}

\theoremstyle{definition}

\theoremstyle{definition}

\theoremstyle{plain}

\theoremstyle{plain}

\theoremstyle{plain}

\theoremstyle{plain}

\title{QISTA-Net: DNN Architecture to Solve $\ell_q$-norm Minimization Problem
and Image Compressed Sensing}

\name{Gang-Xuan Lin, Shih-Wei Hu, and Chun-Shien Lu}
\address{Institute of Information Science, Academia Sinica, Taipei, Taiwan}

\begin{document}

\maketitle

\begin{abstract}

\noindent
In this paper, we reformulate the non-convex $\ell_q$-norm minimization problem
with $q\in(0,1)$ into a 2-step problem,
which consists of one convex and one non-convex subproblems, 
and propose a novel iterative algorithm called
QISTA ($\ell_q$-ISTA) to solve the $\left(\ell_q\right)$-problem.
By taking advantage of deep learning in accelerating optimization algorithms,
together with the speedup strategy that using the momentum from all previous layers in the network, we propose a learning-based method, called QISTA-Net-s, to solve the sparse signal reconstruction problem.
Extensive experimental comparisons demonstrate that 
the QISTA-Net-s yield better reconstruction qualities than state-of-the-art $\ell_1$-norm optimization (plus learning) algorithms even if the original sparse signal is noisy.
On the other hand, based on the network architecture associated with QISTA, with considering the use of convolution layers, we proposed the QISTA-Net-n for solving the image CS problem, and the performance of the reconstruction still outperforms most of the state-of-the-art natural images reconstruction methods.
QISTA-Net-n is designed in unfolding QISTA and adding the convolutional operator as the dictionary.
This makes QISTA-Net-s interpretable.
We provide complete experimental results that QISTA-Net-s and QISTA-Net-n contribute the better reconstruction performance than the competing.

\end{abstract}

\section{Introduction}\label{sec:intro}
\subsection{Background and Problem Definition}
In sparse signal recovery like compressive sensing (CS) \cite{CRT06, Donoho06}, we usually let $x_0\in\mathbb{R}^n$ denote a $k$-sparse signal to be sensed, let $A\in\mathbb{R}^{m\times n}$ represent a sampling matrix, and let $y\in\mathbb{R}^m$ 
be the measurement vector defined as
\begin{eqnarray}\label{CS}
y=Ax_0,
\end{eqnarray}
where $k< m < n$ and $0<\frac{m}{n}<1$ is defined as the measurement rate.
At the decoder, $x_0$ can be recovered based on its sparsity by means of solving $\ell_0$-norm regularization problem:
\begin{eqnarray}\label{L0}
\left(\ell_0\right)\ \ \ \min_x\dfrac{1}{2}\left\|y-Ax\right\|_2^2+\lambda\left\|x\right\|_0,
\end{eqnarray}
where $\lambda>0$ is a regularization parameter.

The $\left(\ell_0\right)$-problem is NP-hard \cite{L0-NP-hard} as it suffers from the non-convexity and discontinuity of objective function so that there is no efficient algorithm to solve its global minima. An effective way to recover the original sparse signal $x_0$ is relaxing the objective function in (\ref{L0}) as the $\ell_1$-norm regularization problem, which is known as ``LASSO'' \cite{SDC03, DDM04}:
\begin{eqnarray}\label{LASSO}
\left(\mbox{LASSO}\right)\ \ \ \min_x\dfrac{1}{2}\left\|y-Ax\right\|_2^2+\lambda\left\|x\right\|_1.
\end{eqnarray}

Nevertheless, considering LASSO cannot recover the original sparse signal under low measurement rates (say $m<3k$) \cite{C07},
$\ell_q$-norm regularization is suggested \cite{C07, CY08}. The (non-convex) $\ell_q$-norm regularization problem has the form
\begin{eqnarray}\label{Lq}
(\ell_q):\ \ \min_x\dfrac{1}{2}\left\|y-Ax\right\|_2^2+\lambda\left\|x\right\|_q^q,
\end{eqnarray}
where $0<q<1$, and $\left\|x\right\|_q=\sum_{i=1}^n\left(\left|x_i\right|^q\right)^{1/q}$ is the
$\ell_q$-quasi-norm (which is usually called $\ell_q$-norm).
In comparison with $\left(\ell_0\right)$-problem and LASSO,
the authors concluded that decreasing
$q$ further decreases the required measurement rate and by less and less as $q$ gets smaller
\cite{C07, CY08}.

It is noted that the discussions regarding
$\left(\ell_q\right)$-problem or effective algorithms in finding its optimal solution are very rare
in the literature. Furthermore, $\left(\ell_q\right)$-problem is also NP-hard \cite{Lq-NP-hard}, and literature review reveals that solving $\left(\ell_q\right)$-problem suffers from non-convexity, leading to a local-non-global optimal solution.
Although it is difficult to find the global optimal solution, under good initial iterative point,
the limit point gained by iterative algorithms converging to a local-non-global optimal solution
is still closer to $x^*_{\ell_0}$ of $\left(\ell_0\right)$-problem than $x^*_{\ell_1}$ of LASSO \cite{ZMWWL17}, where $x^*$ is the global optimal solution.

\subsection{Related Works}

Traditionally, one always approximates the optimal solution to $\left(\ell_0\right)$-problem and LASSO,
by employing proximal gradient descent method (PGD), which is also known as iterative hard-threshold algorithm (IHT) \cite{IHT} and iterative soft-threshold algorithm (ISTA) \cite{DDM04}. However, IHT could obtain better reconstruction quality than ISTA only if the original signal is very sparse ($k/n<5\%$) and/or measurement rate is high ($m/n>50\%$) \cite{WCLQ18, L12}.
The use of PGD to solve $\left(\ell_q\right)$-problem is not popular because there is no closed-form solution to proximal operator associated with its regularization term \cite{WCLQ18}.
Beck and Teboulle speed up ISTA by using Nesterov's acceleration method (insert momentum after gradient descent step), which is known as FISTA \cite{FISTA}, whereas Donoho \emph{et al}. consider an efficient algorithm called AMP (approximate message passing) that incorporates ISTA with Onsager term in measurement residue $y-Ax^t$ \cite{AMP09}.

On the other hand, in solving the $\left(\ell_q\right)$-problem,
Cui \emph{et al}. \cite{ITA} propose to utilize the iterative thresholding (IT) algorithm in finding the global optimal solution of surrogate function.
Xu \emph{et al}. \cite{L12} design a half thresholding algorithm by thresholding representation theory to solve the $\left(\ell_q\right)$-problem when $q=1/2$.
Cao \emph{et al}. \cite{q23} deduce the thresholding formula in \cite{L12} to derive the extension thresholding formula, which can solve the $\left(\ell_q\right)$-problem when $q=2/3$. However, most of the algorithms still suffer from the non-convexity of $\ell_q$-regularized term, leading to a local-non-global optimal solution, though such a solution results in better reconstruct performance than IHT and ISTA, which solve $\left(\ell_0\right)$-problem and LASSO, respectively. Moreover, \cite{L12} and \cite{q23} restrict  the choice of $q$ ($=1/2$ or $=2/3$) and the other aforementioned methods have to tune an appropriate $q$ to get better results, which violates the fact that $q$ should be small \cite{C07}.

Recently, compare with the traditional iterative methods, deep learning-based paradigm
has received considerable attention due to its extremely low complexity and outstanding performance.
By incorporating deep learning models and optimization methods, several deep network-based sparse signal reconstruction methods have been proposed in the literature.

To accelerate the sparse 1D signal reconstruction methods, 
in 2010, Gregor and LeCun \cite{LISTA-LeCun} proposed a learned ISTA method, called LISTA.
Borgerding \emph{et al}. \cite{LVAMP} proposed a learned AMP method, called LAMP.
Based on the theoretical linear convergence of LISTA, Chen \emph{et al}. \cite{LISTA-CPSS}
introduced the necessary conditions for the LISTA, and proposed the LISTA-CP, LISTA-SS, and LISTA-CPSS to improve reconstruction performance with fewer training parameters.
In order to speed up the training process, Ito \emph{et al}. \cite{TISTA} proposed the so-called TISTA
(trainable ISTA) method, which only needs $T+2$ training parameters and has high reconstruction performance, where $T$ is the layer number.

In the reconstruction of natural images, Mousavi \emph{et al}. \cite{SDA} first proposed to apply a stacked denoising auto-encoder (SDA) to learn the representation and to reconstruct natural images from their CS measurements.
Kulkarni \emph{et al}. \cite{ReconNet} further developed a CNN-based method, dubbed ReconNet, to reconstruct the natural images.
To the best of our knowledge, ReconNet probably the first one who solves image CS problem by CNN-based method.
Similar to \cite{ReconNet}, all of the network architectures in MS-CSNet \cite{MS-CSNet},
DR$^2$-Net \cite{DR2-Net}, MSRNet \cite{MSRNet}, CSNet$^+$ \cite{CSNet}, and SCSNet \cite{SCSNet} are heuristic designs in solving CS.
Another category aims to develop learning-based methods designed based on an iterative-based algorithm.
The methods \cite{ISTA-Net} \cite{KC2019} unfolded the parameters of the iterative-based algorithm, within the framework of CNN to design the network architectures.


\subsection{Contributions}
\begin{itemize}
\item[1.] We reformulate $\left(\ell_q\right)$-norm minimization problem into 2-step problem that transfers the difficulty coming from non-convexity to another non-convex optimization problem that can be trivially solved. Then we design an algorithm called QISTA that approximates the optimal solution of $\left(\ell_q\right)$-problem precisely.

\item[2.] QISTA-Net-s is a DNN architecture by unfolding specific parameters in QISTA to accelerate the reconstruction.
After unfolding, we utilize the momentum coming from all previous layers to further speed up the network architecture.
The use of momentum for network architecture is novel in literature.

\item[3.] The performance of QISTA-Net-s in reconstruct the sparse signal is better than state-of-the-art $\ell_1$-norm learning-based methods, even in noisy environments.

\item[4.] In reconstructing the natural images, the proposed method QISTA-Net-n is comparable with state-of-the-art learning-based methods, including the method that specially designed for reconstructing the natural image.
\end{itemize}

\subsection{Organization of This Paper}

The remainder of this paper is organized as follows.
In Sec. \ref{sec:QISTA}, we propose an iterative method, called QISTA, to solve the $\left(\ell_q\right)$-problem (\ref{Lq}).
Based on QISTA, we propose a learning-based method, called QISTA-Net-s, in solving problem (\ref{Lq}) in Sec. \ref{sec:QISTA-Net-s}.
Again, based on QISTA, we propose a learning-based method in solving the image CS problem in Sec. \ref{sec:QISTA-Net-n}.
The comparison with the state-of-the-art methods of QISTA, QISTA-Net-s, and QISTA-Net-n are provided in Sec. \ref{sec:exp QISTA}, Sec. \ref{sec:exp QISTA-Net-s}, and Sec. \ref{sec:exp QISTA-Net-n}, respectively.

\section{Iterative Method for Solving the Sparse Signal Reconstruction Problem: QISTA}\label{sec:QISTA}

In Sec. \ref{sec:2-step-problem}, we first approximate the $\left(\ell_q\right)$-problem (\ref{Lq}) and reformulate it into the $2$-step problem.
We then propose an iterative algorithm, QISTA, for solving the $\left(\ell_q\right)$-problem in Sec. \ref{subsec:QISTA}.
The insight of QISTA is described in Sec. \ref{sec:insight}.

\subsection{Reformulate $\left(\ell_q\right)$-problem as $2$-Step Problem}\label{sec:2-step-problem}

To solve $\left(\ell_q\right)$-problem,
it is first approximated as
\begin{eqnarray}\label{Lq-approx}
\min_x F(x)=\dfrac{1}{2}\left\|y-Ax\right\|_2^2+\lambda\sum_{i=1}^n\dfrac{\left|x_i\right|}{\left(\left|x_i\right|+\varepsilon_i\right)^{1-q}},
\end{eqnarray}
where $\varepsilon_i>0$ for all $i\in[1:n]$. We can see that the objective function
in (\ref{Lq-approx}) is equivalent to the one in (\ref{Lq}) provided $\varepsilon_i=0$ for all $i$, \emph{i.e.},
$$\lim\limits_{\varepsilon_i\rightarrow0^+}\dfrac{\left|x_i\right|}{\left(\left|x_i\right|+\varepsilon_i\right)^{1-q}}=\left|x_i\right|^q.$$
This means the problem (\ref{Lq-approx})
approximates to the $(\ell_q)$-problem (\ref{Lq}) well if $\varepsilon_i$'s are small enough.

Second, we extend $F(x)$ in the problem (\ref{Lq-approx}) into high-dimensional functional $H(x,c)$,
then relax the problem (\ref{Lq-approx}) (in the sense of feasible set from $\mathbb{R}^n$ to
$\mathbb{R}^n\times\mathbb{R}^n$) into
\begin{eqnarray}\label{Gxc}
\min_{x,c}H(x,c)=\dfrac{1}{2}\left\|y-Ax\right\|_2^2+\lambda\sum_{i=1}^n\dfrac{\left|x_i\right|}{\left(\left|c_i\right|+\varepsilon_i\right)^{1-q}}.
\end{eqnarray}

\noindent
We can see that the functional $H(x,c)$ degenerates to $F(x)$ if $c=x$.
Thus, we can reformulate the problem (\ref{Lq-approx}) as a 2-step problem:
\begin{eqnarray}\label{2-step}
\left\{
\begin{array}{r@{\ }l}
\displaystyle\min_x&\displaystyle H(x,\bar{c})=\dfrac{1}{2}\left\|y-Ax\right\|_2^2+\lambda\sum_{i=1}^n\dfrac{\left|x_i\right|}{\left(\left|\bar{c}_i\right|+\varepsilon_i\right)^{1-q}},\\
\displaystyle\min_c&\displaystyle\left|H\left(\bar{x},c\right)-H\left(\bar{x},\bar{x}\right)\right|,
\end{array}
\right.
\end{eqnarray}
where $\bar{x}$ and $\bar{c}$ are optimal solutions to the first problem
(called $x$-subproblem) and the second problem
(called $c$-subproblem), respectively.

\begin{thm}\label{thm:2-step}
If $\left(x^*,c^*\right)$ is an optimal solution pair to 2-step problem (\ref{2-step}), then $x^*$ is an optimal solution to problem (\ref{Lq-approx}), and vice versa.
\end{thm}

\noindent
\textbf{Proof}.
Let $\left(x^*,c^*\right)$ be an optimal solution pair to (\ref{2-step}),
since the optimal value of $c$-subproblem is obviously 0,
we have $H\left(x^*,c^*\right)=H\left(x^*,x^*\right)$, which equals to
$F\left(x^*\right)$ (because $H(x,x)=F(x)$), therefore $x^*$ is an optimal solution to (\ref{Lq-approx}).

On the other hand, let $x^*$ be an optimal solution to (\ref{Lq-approx}), then $c^*=x^*$ is an optimal solution to $c$-subproblem of (\ref{2-step}), whereas $x$-subproblem is exactly equivalent to (\ref{Lq-approx}).
$\hfill\blacksquare$ 

We can see that the $c$-subproblem
has global minimum solution
\begin{eqnarray}\label{c-subproblem}
\bar{c}=\bar{x},
\end{eqnarray}
whereas the $x$-subproblem is in a weighted-LASSO form
\begin{eqnarray}\label{weight_LASSO}
\min_x\dfrac{1}{2}\left\|y-Ax\right\|_2^2+\lambda\sum_{i=1}^n\left|w_ix_i\right|,
\end{eqnarray}
(the weight of $\left|x_i\right|$ is $\frac{\lambda}{\left(\left|\bar{c}_i\right|+\varepsilon_i\right)^{1-q}}$), and thus
the $x$-subproblem can be solved iteratively by proximal gradient descent algorithm \cite{FOMO17} as
\begin{eqnarray}\label{prox_grad}
\left\{
\begin{array}{r@{\ }l}
r^t&=x^t+\beta A^T\left(y-Ax^t\right)\\
x^{t+1}&=\eta\left(r^t;\theta\right),
\end{array}
\right.
\end{eqnarray}
where $\theta_i=\frac{\lambda}{\left(\left|\bar{c}_i\right|+\varepsilon_i\right)^{1-q}}$,
$\forall i$,
(when we solve the $x$-subproblem, $\bar{c}$ is a given
constant vector, so $\theta$ is fixed in each iteration)
$\eta\left(\cdot;\cdot\right)$ is component-wise soft-thresholding operator
$$\eta\left(x;\theta\right)=sign\left(x\right)\cdot\max\left\{0,\left|x\right|-\theta\right\}.$$ 
The iterative process in Eq. (\ref{prox_grad}) consists of two steps, \emph{i.e.}, gradient descent step followed by truncation step.

\subsection{QISTA}\label{subsec:QISTA}

In Eq. (\ref{prox_grad}), the iterative process
is formed by the alternating iteration of two points: $r^t$ and $x^t$.
In the first step (gradient descent step) of Eq. (\ref{prox_grad}),
the step size $\beta$ is always chosen from
$(0,\frac{1}{L_f}]$ \cite{DDM04, FISTA, LVAMP}, where $f(x)=\frac{1}{2}\left\|y-Ax\right\|_2^2$ and $L_f$ is the smallest Lipschitz constant of $\nabla f(x)$. Indeed $L_f=\left\|A\right\|_2^2$, where $\left\|A\right\|_2$ is the spectral norm of $A$.
We can see that $r^t-x^t=\beta A^T\left(y-Ax^t\right)$, 
which implies
\begin{eqnarray*}
\begin{array}{r@{\ }l}
\left\|r^t-x^t\right\|_2^2&=\beta\left\|A^T\left(y-Ax^t\right)\right\|_2^2\\
&\leq\beta\left\|A^T\right\|_2^2\left\|y-Ax^t\right\|_2^2\\
&=\beta\left\|A\right\|_2^2\left\|y-Ax^t\right\|_2^2\\
&\leq\left\|y-Ax^t\right\|_2^2.
\end{array}
\end{eqnarray*}

Since the $x$-subproblem is convex, the proximal gradient descent algorithm
(\ref{prox_grad}) is global-convergence \cite{FISTA}, then we have  $\lim\limits_{t\rightarrow\infty}x^t=\bar{x}$,
and
\begin{eqnarray}\label{rx_eq}
\begin{array}{c}
\lim\limits_{t\rightarrow\infty}\left\|r^t-x^t\right\|_2
\leq\left\|y-A\bar{x}\right\|_2.
\end{array}
\end{eqnarray}

\noindent
On the other hand, if the penalty coefficient $\lambda$ of $x$-subproblem in 2-step problem (\ref{2-step}) is small enough,
the magnitude of the regularization term ($\lambda\sum_{i=1}^n\frac{\left|x_i\right|}{\left(\left|\bar{c}_i\right|+\varepsilon_i\right)^{1-q}}$, the penalty term)
is small relative to the magnitude of the term $f(x)=\frac{1}{2}\left\|y-Ax\right\|_2^2$. This means that the optimal solution $\bar{x}$ to $x$-subproblem leads to a relatively small value $\frac{1}{2}\left\|y-A\bar{x}\right\|_2^2$.
This is to say, if $\lambda$ is small enough,
by Eq. (\ref{rx_eq}), we have
\begin{eqnarray}\label{r_approx_to_x}
\lim\limits_{t\rightarrow\infty}r^t\approx
\lim\limits_{t\rightarrow\infty}x^t=\bar{x}.
\end{eqnarray}

\noindent
In other words, the two points $r^t$ and $x^t$ are close if
$t$ is large enough, $\beta\in(0,\frac{1}{L_f}]$, and $\lambda$ is small enough.
Under the circumstance,
instead of choosing Eq. \ref{c-subproblem} as the solution of the $c$-subproblem in Eq. \ref{2-step}, the approximation 
\begin{eqnarray}\label{c-subproblem-2}
\bar{c}\approx r^t,
\end{eqnarray}
where $r^t$ is the iterative point of the proximal gradient descent algorithm
that solves the $x$-subproblem, was adopted.
The benefits of using the approximate optimal solution instead of the exact optimal solution
for $c$-subproblem, will be discussed in Sec. \ref{sec:insight}.

At last, we summarized the alternative iteration for solving the 2-step problem (\ref{2-step}) as follows:
\begin{eqnarray}\label{QISTA}
\left\{
\begin{array}{r@{\ }l}
r^t&=x^{t-1}+\beta A^T\left(y-Ax^{t-1}\right),\\
c^t&=r^t,\\
x_i^t&=\eta
\left(r_i^t;\dfrac{\lambda}{\left(\left|c_i^t\right|+\varepsilon_i\right)^{1-q}}\right),\ \forall i
\end{array}
\right.
\end{eqnarray}
which is equivalent to
\begin{eqnarray}\label{QISTA-merge}
(\mbox{QISTA}):\ \ \left\{
\begin{array}{r@{\ }l}
r^t&=x^{t-1}+\beta A^T\left(y-Ax^{t-1}\right),\\
x_i^t&=\eta
\left(r_i^t;\dfrac{\lambda}{\left(\left|r_i^t\right|+\varepsilon_i\right)^{1-q}}\right),\ \forall i.
\end{array}
\right.
\end{eqnarray}
We name it QISTA ($\ell_q$-ISTA).

\begin{algorithm}[h]
  \caption{QISTA}
  \begin{algorithmic}[1]
  \State Set parameters $\beta$, $\lambda$, TOL;
  \State Initial $x^{\mbox{\footnotesize init}}=x^0=x^{-1}\in\mathbb{R}^n$;
    \Repeat
      \State $r^t=x^{t-1}+\beta A^T\left(y-Ax^{t-1}\right)$;
      \State $x_i^t=\eta
\left(r_i^t;\dfrac{\lambda}{\left(\left|r_i^t\right|+\varepsilon_i\right)^{1-q}}\right),\ \forall\ i\in[1:n]$;
    \Until{$\left(\left\|x^t-x^{t-1}\right\|_2<TOL\right)$}
  \end{algorithmic}
  \label{alg:QISTA}
\end{algorithm}

\noindent

\subsection{Insight of QISTA}\label{sec:insight}

\begin{figure}[!hbpt]
\begin{center}
\includegraphics[width=\linewidth]{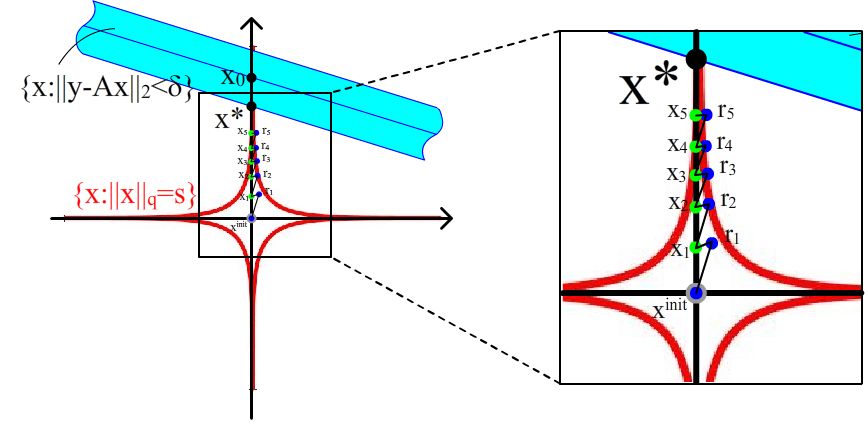}
\caption{The iterative process of QISTA. The water-colored region is the set
$\left\{x:\left\|y-Ax\right\|_2<\delta\right\}$, where $\delta$ is a constant related to $\lambda$.
The red line is the contour line $\left\{x:\left\|x\right\|_q=s\right\}$ for a constant $s$.
$x_0$ is the ground-truth, whereas $x^*$ is the optimal solution to $\left(\ell_q\right)$-problem.}
\label{fig:insight}
\end{center}
\end{figure}

The iterative process in Algorithm \ref{alg:QISTA} consists of two steps, \emph{i.e.}, the gradient descent step (step 4), followed by the truncation step (step 5).
The gradient descent step updates the point by moving the current iterative point $x^{t-1}$, along the direction $A^T\left(y-Ax^{t-1}\right)$ which perpendicular to the null
space of $A$, with the step size $\beta$, to the updated point $r^t$ (as shown in Fig. \ref{fig:insight}, moving the green point $x^{t-1}$ to the blue point $r^t$).
In the truncation step, the truncated parameter
$$\theta_{r^t}(i)=\frac{\lambda}{\left(\left|r_i^t\right|+\varepsilon_i\right)^{1-q}}$$
is determined by $\left|r_i^t\right|$ ($\lambda$, $\varepsilon_i$, and $q$ are constants).
This says that if $\left|r_i^t\right|$ is non-zero, or larger than the other components $\left|r_j^t\right|$
(which we guess that the index $i$ may in the support set of $x_0$),
then $\theta_{r^t}(i)$ is relatively small.
Conversely, if $\left|r_i^t\right|$ is zero, or is a relatively small number (which we guess that
the $i^{\mbox{\footnotesize th}}$ component of $x_0$ is zero),
then $\theta_{r^t}(i)$ is relatively large.

The input of the truncation step, $r^t$, is truncated by $\theta_{r^t}$, componentwise.
We can see that if $\left|r_i^t\right|$ is a large number, the $\theta_{r^t}(i)$ is almost zero,
and then the truncate operator $\eta\left(r_i^t;\theta_{r^t}(i)\right)$ will preserve the value of $r_i^t$.
Conversely, if $\left|r_i^t\right|$ is a small number,
the truncate operator $\eta\left(r_i^t;\theta_{r^t}(i)\right)$ will decrease the value of $\left|r_i^t\right|$. 
As shown in Fig. \ref{fig:insight}, the truncation step in QISTA updates the point $r^t$ (blue points)
by moving it,
along the direction perpendicular to the set
$\{x:\left\|x\right\|_q=\left\|r^t\right\|_q\}$ (is a contour line $\left\{x:\left\|x\right\|_q=s\right\}$ for some constant $s$) approximately,
to the point $x^t$ (green points).
As the iteration progresses, the $x^t=\eta\left(r^t;\theta_{r^t}\right)$ will gradually approach the optimal solution to $\left(\ell_q\right)$-problem.
Remark that the truncate parameter $\theta_{r^t}$ adapts to the value of $r^t$, componentwise, instead of applying the same truncation value to every component, as of ISTA.
This indicates why QISTA is better than ISTA.

\subsection{Remarks}
Basically, in QISTA, we first approximate the $\left(\ell_q\right)$-problem by (\ref{Lq-approx}), and then reformulate it into the $2$-step problem (\ref{2-step}).
Since the objective function $F(x)$ in (\ref{Lq-approx}) is non-convex, it is difficult to attain the global minima; whereas the functional $H(x,c)$ in (\ref{2-step}) is convex in $x$ for any given $c$, but non-convex in $c$ for any given $x$.
However, since the $x$-subproblem is convex, the proximal gradient descent algorithm
(\ref{prox_grad}) to find its optimal solution is global-convergence under mild parameter setting \cite{FISTA}.
On the other hand, although the $c$-subproblem is non-convex,
the non-convexity can be avoided because there is a trivial global minimum solution, \emph{i.e.}, $\bar{c}=\bar{x}$.

\section{Learning-Based Method for Solving the Sparse Signal Reconstruction Problem: QISTA-Net-s}\label{sec:QISTA-Net-s}

Similar to \cite{LISTA-LeCun, LVAMP, LISTA-CPSS}, to  accelerate QISTA, we design a deep neural network (DNN) architecture by unfolding the parameters $\beta A^T,
\lambda, $ and $\varepsilon$, into $\mathcal{A}^t$, $\lambda^t$, and $\mathcal{E}^t$, respectively. The network architecture then has the form
\begin{eqnarray}\label{eq:QISTA-unfold}
\left\{
\begin{array}{r@{\ }l}
r^t&=x^{t-1}+\mathcal{A}^t\left(y-Ax^{t-1}\right),\\
x_i^t&=\eta
\left(r_i^t;\dfrac{\lambda^t}{\left(\left|r_i^t\right|+\mathcal{E}_i^t\right)^{1-q}}\right),\ \forall\ i\in[1:n],
\end{array}
\right.
\end{eqnarray}

\noindent
where $t$ stands for the $t^{\mbox{\footnotesize th}}$ layer of
the network.

The insight of (\ref{eq:QISTA-unfold}) is that,
the first equation is corresponding to the gradient descent step,
by letting $\beta A^T$ be a training parameter, the NN architecture is trying to
modify the descent direction and find the appropriate step size simultaneously.
Besides, $\lambda^t$ in the second equation is to learn the appropriate truncated rate with respect to
the current iterative point $r^t$; and, $\mathcal{E}^t$ is to control the similarity
between approximated ($\ell_q$)-problem (\ref{Lq-approx}) and 
the original ($\ell_q$)-problem (\ref{Lq}).

In FISTA \cite{FISTA}, the gradient descent step considers the previous iterative direction called momentum.
And, in AMP \cite{AMP09}, the residue in the measurement domain considers the previous iterative residue called the Onsager term.
Inspired by FISTA and AMP, we further accelerate the network in (\ref{eq:QISTA-unfold})
by adding the ``momentum'' that coming from the descent direction of
all previous layers with an appropriate weight. The resultant is shown in Algorithm
\ref{alg:QISTA-Net-s}, which is called QISTA-Net-s (QISTA in Network model for Sparse signal reconstruction).

\begin{algorithm}[h]
  \caption{QISTA-Net-s}
  \begin{algorithmic}[1]
    \For{$t=1$ to $T$}
      \State $D^t=\mathcal{A}^t\left(y-Ax^{t-1}\right)$;
      \State $r^t=x^{t-1}+\sum_{j=1}^tD^j$;
      \State $D^j=\frac{\gamma}{m}\cdot D^j,
\ \forall j\in[1:t]$;
      \State $x_i^t=\eta
\left(r_i^t;\dfrac{\lambda^t}{\left(\left|r_i^t\right|+\mathcal{E}_i^t\right)^{1-q}}\right),\ \forall\ i\in[1:n]$;
    \EndFor
  \end{algorithmic}
  \label{alg:QISTA-Net-s}
\end{algorithm}

\noindent
The learning parameters in Algorithm \ref{alg:QISTA-Net-s} are $\left\{\mathcal{A}^t,\lambda^t,\mathcal{E}^t\right\}_{t=1}^T$, and the loss function is the MSE-loss:
\begin{eqnarray}\label{eq:MSE loss}
\mathcal{L}_{\mbox{\footnotesize MSE}}=\frac{1}{n}\left\|x_0-x^T\right\|_2^2.
\end{eqnarray}
where $x^T$ is the output of the algorithm.

In Algorithm \ref{alg:QISTA-Net-s},
$D^t$ in Step 2 is the descent direction of the current layer $t$,
$\sum_{j=1}^{t-1}D^j$ in Step 3 is the momentum consisting of the descent direction of previous layers, and
Step 4 controls the effect of momentum appropriately.
However, unlike the acceleration of traditional iterative methods such as FISTA \cite{FISTA}, the improvement effect in QISTA-Net-s can only draw empirical conclusions without being able to conduct mathematical analysis.

Remark that the first equation in (\ref{eq:QISTA-unfold})
is corresponding to the gradient descent step (as in Step 4 of Algorithm \ref{alg:QISTA}),
which unfold both $A$ and $\beta A^T$ to be learning parameters is
commonly adopted in LISTA \cite{LISTA-LeCun}, LAMP \cite{LVAMP}, and other
learning-based methods.
According to Theorem 1 in \cite{LISTA-CPSS}, we only set $\beta A^T$ to be a learning parameter and keep $A$ as the original matrix to reduce the training time without loss of reconstruction performance. 

\section{Learning-Based Method for Recunstructing the Natural Images:
QISTA-Net-n}\label{sec:QISTA-Net-n}

In Sec. \ref{sec:QISTA-Net-s}, we proposed the QISTA-Net-s for the sparse 1D signal reconstruction problem.
Here we address the other critical issue called natural image reconstruction problem. 

In this section, we consider the natural image $X_0\in\mathbb{R}^{n_1\times n_2}$, where $n_1\times n_2$ is image size, and its vector representation $x_0\in\mathbb{R}^{n}$, where $n=n_1\cdot n_2$. 
The natural image is, in general, a non-sparse signal, and it always exists a sparse representation in a specific domain.
The traditional optimization method
usually reconstruct the original image $x_0$ by solving the following $\ell_1$-norm regularization problem in LASSO form:
\begin{eqnarray}\label{L1-dictionary}
\min_x\dfrac{1}{2}\left\|y-Ax\right\|_2^2+\lambda\left\|\Psi x\right\|_1,
\end{eqnarray}
where $A\in\mathbb{R}^{m\times n}$ is the sensing matrix,
and $\Psi\in\mathbb{R}^{n\times n}$ is dictionary
that represent $x_0$ in sparse coefficient with
respect to the specific domain.
In this section, we consider the $\ell_q$-norm regularization
problem in the form:
\begin{eqnarray}\label{Lq-dictionary}
\min_x\dfrac{1}{2}\left\|y-Ax\right\|_2^2+\lambda\left\|\Psi x\right\|_q^q,
\end{eqnarray}
where $0<q<1$.

Similar to the process from Eq. (\ref{Lq-approx}) to Eq. (\ref{2-step}) in Sec. \ref{sec:2-step-problem},
we can reformulate the problem (\ref{Lq-dictionary}) as a 2-step problem:
\begin{eqnarray}\label{2-step-dictionary}
\left\{
\begin{array}{r@{\ }l}
\displaystyle\min_x&\displaystyle H(x,\bar{c})=\dfrac{1}{2}\left\|y-Ax\right\|_2^2+\lambda\sum_{i=1}^n\dfrac{\left|\left(\Psi x\right)_i\right|}{\left(\left|\bar{c}_i\right|+\varepsilon_i\right)^{1-q}},\\
\displaystyle\min_c&\displaystyle\left|H\left(\bar{x},c\right)-H\left(\bar{x},\Psi\bar{x}\right)\right|,
\end{array}
\right.
\end{eqnarray}
where $\varepsilon_i>0$ for all $i\in[1:n]$, and $\bar{x}$ and $\bar{c}$ are optimal solutions to the
$x$-subproblem and the $c$-subproblem, respectively.

\subsection{Generalized proximal operator}\label{sec:general_proximal_operator}

In this subsection, we aim to design an explicit iterative process
that solves the $x$-subproblem in (\ref{2-step-dictionary}).
In the $x$-subproblem, 
we can see that the regularization term of the
objective function is in the form of the composite function
$\left\|\Psi(x)\right\|_{1,w}=\left(\left\|\cdot\right\|_{1,w}\circ\Psi\right)(x)$,
where $\left\|x\right\|_{1,w}=\sum_{i=1}^nw_i\left|x_i\right|$
with $w_i=\frac{\lambda}{\left(\left|\bar{c}_i\right|+\varepsilon_i\right)^{1-q}}$.
Therefore the proximal gradient descent algorithm \cite{FOMO17} for solving the $x$-subproblem
has the form
\begin{eqnarray}\label{prox_grad_dictionary}
\left\{
\begin{array}{r@{\ }l}
r^t&=x^{t-1}+\beta A^T\left(y-Ax^{t-1}\right),\\
x^t&=\mbox{prox}_{\left\|\Psi(\cdot)\right\|_{1,w}}\left(r^t\right).
\end{array}
\right.
\end{eqnarray}

\noindent
Different from the problem (\ref{weight_LASSO}), 
there is no useful calculus rule for computing the proximal operator of
a composite function $\left\|\Psi\left(\cdot\right)\right\|_{1,w}$ so that Eqs. (\ref{prox_grad_dictionary}) 
cannot be explicitly written as the iterative process similar to Eqs. (\ref{prox_grad}).
To address the circumstance, we introduce the generalized proximal operator by the following theorem:
\begin{thm}\label{prox thm}\cite{FOMO17}
Let $g:\mathbb{R}^n\rightarrow(-\infty, \infty]$ be a proper closed convex function, and let $f(x) = g(\mathcal{A}(x)+b)$, where $b\in\mathbb{R}^n$ and $\mathcal{A}:\mathbb{R}^{\hat{n}}\rightarrow\mathbb{R}^n$ is a linear transformation satisfying $\mathcal{A}\circ\mathcal{A}^T=\gamma\cdot I_n$ for some constant $\gamma >0$. Then for any $x\in\mathbb{R}^{\hat{n}}$,
$$\mathrm{prox}_f(x) = x+\frac{1}{\gamma}\mathcal{A}^T\left( \mathrm{prox}_{\gamma g}(\mathcal{A}(x)+b)-(\mathcal{A}(x)+b) \right). $$
\end{thm}

As decribed in theorem \ref{prox thm}, we can see that if the dictionary $\Psi$ is linear and satisfies a certain orthogonality condition, 
the solution to the proximal operator of the regularized term
$\lambda\sum_{i=1}^n\frac{\left|\left(\Psi x\right)_i\right|}{\left(\left|\bar{c}_i\right|+\varepsilon_i\right)^{1-q}}$
in $x$-subproblem can be found.

After replacing $g(x)$ and $\mathcal{A}(x)$ in Theorem \ref{prox thm}
by $\left\|\cdot\right\|_{1,w}$ and $\Psi(x)$, respectively,
the solution to the proximal operator in Eq. (\ref{prox_grad_dictionary}) becomes
\begin{eqnarray}\label{closed-form-sol}
\begin{array}{ll}
&\mbox{prox}_{\left\|\Psi(\cdot)\right\|_{1,w}}\left(r^t\right)\\[+10pt]
=&r^t
+ \frac{1}{\gamma_c}\Psi^{T}
\left(\eta\left(\Psi\left(r^t\right);\gamma_c\right) -\Psi(r^t)\right),
\end{array}
\end{eqnarray}
where $\gamma_c=\frac{\lambda}{\left(\left|\bar{c}\right|+\varepsilon\right)^{1-q}}$, and
we call Eq. (\ref{closed-form-sol}) a generalized proximal operator.

Similar to Eq. (\ref{prox_grad}),
the proximal gradient descent algorithm for solving $x$-subproblem
in (\ref{2-step-dictionary})
can be written in the iterative process as
\begin{eqnarray}\label{prox_grad_dictionary_iter}
\left\{
\begin{array}{r@{\ }l}
r^t&=x^{t-1}+\beta A^T\left(y-Ax^{t-1}\right)\\
x^t&=r^t
+ \frac{1}{\gamma_c}\Psi^{T}
\left(\eta\left(\Psi\left(r^t\right);\gamma_c\right) -\Psi(r^t)\right).
\end{array}
\right.
\end{eqnarray}

On the other hand, $\Psi$ in problem (\ref{Lq-dictionary}) plays the role of a dictionary
that, in general, represent the non-sparse signal as a sparse coefficient with respect to
the specific domain.
However, the classic fixed domain (\emph{e.g.} DCT, DFT, wavelet, gradient domain, etc.) usually result in poor reconstruction performance. Besides, $\Psi$
is generally treated to be an over-complete dictionary (\emph{i.e.}, $\Psi\in\mathbb{R}^{N\times n}$ with $N > n$) to seek better performance.

However, as $N>n$, the assumption $\Psi\circ\Psi^T=\gamma_c I_N$ in Theorem \ref{prox thm} is not satisfied at all.
Thus, it is required to choose a $\Psi^{\dag}$ that satisfies $\Psi\circ\Psi^{\dag}\approx\gamma_c I_N$.
We can see that the left inverse of $\Psi$ always exists, \emph{i.e.},
$\tilde{\Psi}=\left(\Psi^T\circ\Psi\right)^{-1}\circ\Psi^T$ satisfy $\tilde{\Psi}\circ\Psi=I_n$, 
due to the fact that $N>n$.
Then we have
\begin{eqnarray}\label{eq:left inverse}
\begin{array}{r@{\ }l}
\Psi^{\dag}&=I_{n} \circ \Psi^{\dag}\\
&=\left(\tilde{\Psi}\circ\Psi\right)\circ\Psi^{\dag}\\
&=\tilde{\Psi}\circ\left(\Psi\circ\Psi^{\dag}\right)\\
&\approx\tilde{\Psi}\circ\gamma_c I_N\\
&=\gamma_c\tilde{\Psi}.
\end{array}
\end{eqnarray}

\noindent
Therefore, the assumption in Theorem \ref{prox thm} can be relaxed as
$\Psi^T=\bar{\gamma}\tilde{\Psi}$, where $\bar{\gamma}$ is a constant.
Thus, the solution to the proximal operator in
Eq. (\ref{prox_grad_dictionary}) can be approximated as
\begin{equation}\label{general closed-form x}
x^t =r^t + \bar{\gamma}\tilde{\Psi}\left(\eta(\Psi(r^t);\gamma_c) -\Psi(r^t)\right).
\end{equation}

\subsection{Convolutional Dictionary}\label{sec:convolutional dictionary}

Inspired by the representation power of CNN \cite{DLHT14} and its universal approximation property \cite{HSW89},
and the approach to sparsify natural images in \cite{ISTA-Net},
the dictionary $\Psi$ is adopted in the form 
as
\begin{eqnarray}\label{eq:Psi}
\Psi=\mathcal{C}_3\left(\mbox{ReLU}\left(\mathcal{C}_2\left(\mbox{ReLU}\left(\mathcal{C}_1(x)\right)\right)\right)\right),
\end{eqnarray}
where $\mathcal{C}_1$, $\mathcal{C}_2$, and $\mathcal{C}_3$ are convolution operators, and ReLU is a rectified linear unit, and we call this form as an convolutional dictionary.

On the other hand, to exhibit a ``left-inverse'' structure of $\Psi$, $\tilde{\Psi}$ is also adopted in the form of the convolutional dictionary as
\begin{eqnarray}\label{eq:PsiDag}
\tilde{\Psi}=\mathcal{C}_6\left(\mbox{ReLU}\left(\mathcal{C}_5\left(\mbox{ReLU}\left(\mathcal{C}_4(x)\right)\right)\right)\right),
\end{eqnarray}
where
$\mathcal{C}_4$, $\mathcal{C}_5$, and $\mathcal{C}_6$ are convolution operators.
Based on the relaxation that $\Psi^T=\bar{\gamma}\tilde{\Psi}$,
we present a suitable loss function appropriately to ensure the left-inverse relation
between $\Psi$ and $\tilde{\Psi}$.
The loss function will be described in Sec. \ref{sec:loss-n}.

\subsection{QISTA-Net-n}\label{subsec:QISTA-Net-n}

After determining the dictionaries,
the iterative process (\ref{prox_grad_dictionary_iter})
becomes:
\begin{eqnarray}\label{prox_grad_dictionary_Psi}
\left\{
\begin{array}{r@{\ }l}
r^t&=x^{t-1}+\beta A^T\left(y-Ax^{t-1}\right)\\
x^t &=r^t + \bar{\gamma}\tilde{\Psi}\left(\eta(\Psi(r^t);\gamma_c) -\Psi(r^t)\right),
\end{array}
\right.
\end{eqnarray}
where $\Psi$ and $\tilde{\Psi}$ are as shown in
Eqs. (\ref{eq:Psi}) and (\ref{eq:PsiDag}).

Since Eq. (\ref{prox_grad_dictionary_Psi}) is the iterative process
that solves the $x$-subproblem in 2-step problem (\ref{2-step-dictionary}),
and, similar to Sec. \ref{subsec:QISTA},
we also approximate the solution
to the $c$-subproblem by $\bar{c}\approx \Psi\left(r^t\right)$ (as in Eq. (\ref{c-subproblem-2}))
instead of $\Psi\left(x^t\right)$.
Therefore, similar to Eq. (\ref{QISTA}) and Eq. (\ref{QISTA-merge}),
we also propose an iterative process that solves the 2-step problem
(\ref{2-step-dictionary}) as
\begin{eqnarray}\label{prox_grad_dictionary_Psi-2}
\left\{
\begin{array}{r@{\ }l}
r^t&=x^{t-1}+\beta A^T\left(y-Ax^{t-1}\right)\\
x^t &=r^t + \bar{\gamma}\tilde{\Psi}\left(\eta(\Psi(r^t);\gamma_r) -\Psi(r^t)\right),
\end{array}
\right.
\end{eqnarray}
where $\gamma_r=\frac{\lambda}{\left(\left|\Psi\left(r^t\right)\right|+\varepsilon\right)^{1-q}}$.

Similar in Sec. \ref{sec:QISTA-Net-s}, we design a DNN architecture
by unfolding the parameters
$\beta$, $A^T$, $A$, $\bar{\gamma}$, $\tilde{\Psi}$,
$\Psi$, and $\lambda$, into
$\beta^t$, $\mathcal{B}$, $\mathcal{A}$, ${\alpha}^t$, $\tilde{\Psi}^t$,
$\Psi^t$, and $\lambda^t$, respectively.
The network architecture then has the form
\begin{eqnarray}\label{QISTA-Net-n}
\left\{
\begin{array}{r@{\ }l}
r^t&=x^{t-1}+\beta^t \mathcal{B}\left(y-\mathcal{A}x^{t-1}\right)\\
x^t &=r^t + \alpha^t\tilde{\Psi}^t\left(\eta(\Psi^t(r^t);\gamma_r^t) -\Psi^t(r^t)\right),
\end{array}
\right.
\end{eqnarray}
where $\gamma_r^t=\frac{\lambda^t}{\left(\left|\Psi^t\left(r^t\right)\right|+\varepsilon\right)^{1-q}}$.

Remark that the fully connected (FC) layers $\mathcal{B}$
and $\mathcal{A}$ are
common used in every layers, and the other training parameters
are trained independently with respect to the layer.
Besides, the structures of $\tilde{\Psi}^t$ and $\Psi^t$ are
as shown in Eqs. (\ref{eq:Psi}) and (\ref{eq:PsiDag}),
therefore the training parameters is indeed the convolution
operators $\mathcal{C}_i$, $i=1,2,\cdots,6$.

In (\ref{QISTA-Net-n}), the first step consists of
matrix and vector multiplications and affine transformations,
that is, the calculation in the first step is affine-linear,
and the current iterative point is in the time domain.
However, the second step consists of non-linear operators ReLU,
and convolution operators. Specifically, the input $r^t$ of this step
is in the time domain, whereas we hope the convolutional operator
$\Psi^t$, which play
the role of the dictionary, transform $r^t$ into a specific domain
which should represent $r^t$ as a sparse vector.
Therefore, before the $\Psi^t$ act on $r^t$,
we consider an extra convolution operator to transform the $r^t$
in the time domain into the convolution domain,
to helping $\Psi^t$ more
easily to learn a good sparse representation dictionary.
That is, we act the convolution operator $\mathcal{C}_0^t$
before the $\Psi^t$, then $\Psi^t\left(r^t\right)$ in Eq. (\ref{QISTA-Net-n})
becomes
$$\Psi^t\left(r^t\right)
\leftarrow
\Psi^t\left(\mathcal{C}_0^t\left(r^t\right)\right).$$

\noindent
Also, after the $\tilde{\Psi}^t$ operated, we also consider
an extra convolution operator $\mathcal{C}_7^t$ to transform the
$\tilde{\Psi}^t\left(s^t\right)$, where
$s^t=\eta(\Psi^t(r^t);\gamma^t) -\Psi^t(r^t)$,
in the specific domain back into the time domain.
That is, $\tilde{\Psi}^t\left(s^t\right)$ becomes
$$\tilde{\Psi}^t\left(s^t\right)
\leftarrow\mathcal{C}_7^t\left(\tilde{\Psi}^t\left(s^t\right)\right).$$

\noindent
With the extra convolution transformations, the training parameters
$$\left\{\beta^t, \mathcal{B}, \mathcal{A}, \alpha^t, \lambda^t,
\mbox{ and } \mathcal{C}_i^t, i=0, 1, \cdots, 7\right\}.$$
And the forwarding process of QISTA-Net-n
(QISTA-Net for reconstructing the Natural image)
is shown in Algorithm \ref{alg:QISTA-Net-n}.

\begin{algorithm}[h]
  \caption{QISTA-Net-n}
  \begin{algorithmic}[1]
    \For{$t=1$ to $T$}
      \State $r^t=x^{t-1}+\beta^t \mathcal{B}\left(y-\mathcal{A}x^{t-1}\right)$;
      \State $\gamma^t_i=\frac{\lambda^t}{\left(\left|\Psi^t(\mathcal{C}_0^t\left(r^t\right))_i\right|+\varepsilon_i\right)^{1-q}},
      \ \forall\ i\in[1:n];$
      \State $x^t =r^t + \alpha^t\mathcal{C}_7^t\left(\tilde{\Psi}^t\left(\eta(\Psi^t(\mathcal{C}_0^t\left(r^t\right));\gamma^t) -\Psi^t(\mathcal{C}_0^t\left(r^t\right))\right)\right);$
    \EndFor
  \end{algorithmic}
  \label{alg:QISTA-Net-n}
\end{algorithm}

In the forwarding process of QISTA-Net-n,
step 2 is similar to as step 2 plus step 3 in QISTA-Net-s (Algorithm
\ref{alg:QISTA-Net-s}).
The difference is that, in QISTA-Net-s, the $\beta A^T$
is setting to be the training variables
$\mathcal{A}^t$ (independent among each layer) and keep $A$ fixed
as a constant matrix,
whereas in QISTA-Net-n, all of the $\beta$, $A^T$, and $A$ are
setting to be the training variables
$\beta^t$, $\mathcal{B}$, and $\mathcal{A}$, respectively. Note that $\mathcal{A}$ and
$\mathcal{B}$ in QISTA-Net-n are common used in each layer,
and $\beta^t$ is independent among each layer.

\subsection{Loss Function}\label{sec:loss-n}

In general, the loss function of the learning-based method is to consider MSE-loss
as in Eq. (\ref{eq:MSE loss}).
However, in Sec. \ref{sec:general_proximal_operator}, we relaxed the $\Psi^T$ as $\bar{\gamma}\Psi^{\dag}$, therefore we restrict the left-inverse relation between $\Psi$ and $\tilde{\Psi}$ \emph{in each layer $t=1, 2, \cdots, T$}
as
$$\mathcal{L}_{\mbox{\footnotesize aux}}=\sum_{t=1}^T\left\|\tilde{\Psi}^t\circ\Psi^t-I\right\|_2^2,$$
which is implemented by
$$\mathcal{L}_{\mbox{\footnotesize aux}}=\sum_{t=1}^T\left\|\tilde{\Psi}^t\left(\Psi^t\left(w^t\right)\right)-w^t\right\|_2^2,$$
where $w^t=\mathcal{C}_0^t\left(r^t\right)$ in the $t^{\mbox{\footnotesize th}}$-layer.
Together with $\mathcal{L}_{\mbox{\footnotesize MSE}}$, the loss function of QISTA-Net-n is designed as follows:
$$\mathcal{L}=\mathcal{L}_{\mbox{\footnotesize MSE}}
+\delta \mathcal{L}_{\mbox{\footnotesize aux}},$$
where $\delta>0$ is a constant.

\section{Experimental Results of Iterative Method for
Solving the Sparse Signal Reconstruction Problem}\label{sec:exp QISTA}

In this section, we show the reconstruction performance of QISTA
in Algorithm \ref{alg:QISTA} that solving the
sparse signal reconstruction problem in the conventional
iterative optimization style and compare the performance
with state-of-the-art iterative sparse signal reconstruction methods.\footnote{All of our implementation codes
(including of in Sec. \ref{sec:exp QISTA-Net-s} and \ref{sec:exp QISTA-Net-n})
can be downloaded from
https://github.com/spybeiman/QISTA-Net/}

\subsection{Parameter Setting}\label{sec:Parameter Setting QISTA}

For fair comparison, we followed the same setting as in
\cite{ITA} that the
problem dimensions were $n=1024$ and $m=256$, and the ground-truth
$x_0\in\mathbb{R}^n$ was $k$-sparse signal, where the non-zero entries 
followed i.i.d. Gaussian distribution $\mathcal{N}(0,1)$.
For the sensing matrix $A\in\mathbb{R}^{m\times n}$,
its entries $A_{i,j}$'s followed i.i.d. Gaussian distribution
$\mathcal{N}(0,1)$ (without column normalization).

The parameters in QISTA were $\beta=\frac{1}{\left\|A\right\|_2^2}$,
where $\left\|A\right\|_2$ is the spectral norm of $A$,
$\lambda=10^{-4}\cdot\beta$, $q = 0.05$, and $\varepsilon=\textbf{1}_n$,
where $\textbf{1}_n$ is a vector in $\mathbb{R}^n$
with each component being equal to 1.

\subsection{Performance Comparison of Reconstructing
The Synthesize Data}\label{sec:Performance Comparison QISTA}

\begin{figure}[!hbpt]
\begin{center}
\includegraphics[width=\linewidth]{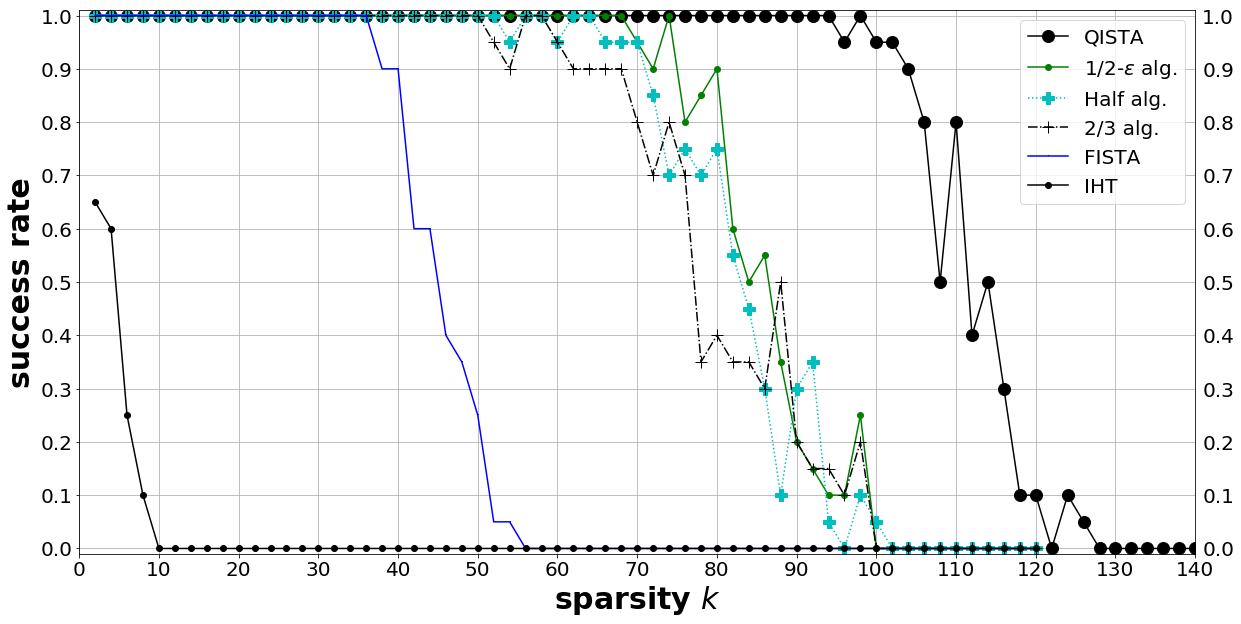}
\caption{The success rates of QISTA,
$1/2-\epsilon$ algorithm,
half thresholding algorithm,
2/3 algorithm,
FISTA,
and IHT in reconstructing the exactly $k$-sparse ground-truth with $n=1024, m=256$, and various $k$.}
\label{compare_iterative_methods}
\end{center}
\end{figure}

We compare the proposed method, QISTA, with traditional iterative methods IHT \cite{IHT}, FISTA \cite{FISTA}
(ISTA \cite{DDM04} was not included because both ISTA and FISTA have exactly the same
reconstruction performance), half thresholding algorithm \cite{L12}, 2/3 algorithm \cite{q23}, 
and $1/2-\epsilon$ algorithm \cite{ITA}.
The criterion, declaring a successful perfect reconstruction of the ground-truth if the relative error
RE$=\frac{\left\|x^*-x_0\right\|_2}{\left\|x_0\right\|_2}\leq10^{-4}$ holds \cite{ITA}, was adopted.
In Fig. \ref{compare_iterative_methods}, the results were shown in terms of the success rate averaged at $20$ tests vs. sparsity $k$.
We can see that QISTA can perfect reconstruct the ground-truth until $k$ is 94 (in this case, $3k=282>m$) but LASSO fails to recover the ground-truth \cite{C07}.
The $1/2-\varepsilon$ algorithm \cite{ITA} perfectly reconstructs the ground-truth
until $k$ is around 68.
The reconstruction performance of QISTA outperforms these state-of-the-art $\ell_q$-based algorithms in terms of success rate.
Remark that AMP \cite{AMP09} was not included for comparison because, under the setting of sensing matrix $A$ without column normalization, it does not converge.

\section{Experimental Results of Deep Learning-Based
Method for Solving the Sparse Signal Reconstruction
Problem}\label{sec:exp QISTA-Net-s}

In this section, we show the reconstruction performance of
QISTA-Net-s in Algorithm \ref{alg:QISTA-Net-s}
that solving the sparse signal reconstruction problem
in the deep learning-based style and compare the performance
with state-of-the-art deep leaning-based sparse signal reconstruction methods.
The QISTA-Net-s were conducted on the device NVIDIA Geforce
GTX 1060 GPU, on the platform Python 3.6 with
Pytorch version 0.4.1.

\subsection{Parameter Setting}\label{sec:Parameter Setting QISTA-Net-s}

For fair comparison, we followed the same setting as in
\cite{LVAMP, LISTA-CPSS, ALISTA} that
the problem dimensions were $n=500$, $m=250$.
The entries of the ground-truth
$x_0\in\mathbb{R}^n$ (which is a $k$-sparse signal)
followed i.i.d. Gaussian distribution $\mathcal{N}(0,1)$
with probability $10\%$ (that is $x_0$ is Bernoulli-Gaussian with $k\approx n\times10\%=50$).
For the sensing matrix $A\in\mathbb{R}^{m\times n}$,
its entries $A_{i,j}$'s followed i.i.d. Gaussian distribution
with column normalization $\mathcal{N}(0,\frac{1}{m})$.

The constant parameters in QISTA-Net-s were 
$\beta=\frac{1}{\left\|A\right\|_2^2}$, $q=0.05$,
and $\gamma=0.1$.
The training parameters of QISTA-Net-s were initialized as
$\lambda^t=10^{-4}\cdot\beta$, $\mathcal{A}^t=\beta A^T$, and $\mathcal{E}^t=0.1\cdot\textbf{1}_n$ for each layer $t$.
In addition, since $\mathcal{E}_i^t$ plays the same role with $\varepsilon_i$ in Eq. (\ref{Lq-approx}), and the value of $\mathcal{E}_i^t$ may be negative after doing back-propagation,
we further restrict $\mathcal{E}_i^t$ after each back-propagation to remain positive by letting $\mathcal{E}_i^t=\max\left\{\mathcal{E}_i^t, 0.1\right\}$.

\subsection{Performance Comparison of Reconstructing
The Synthesize Data}\label{sec:Performance Comparison QISTA-Net-s}

\begin{figure}[!h]
\begin{center}
\includegraphics[width=\linewidth]{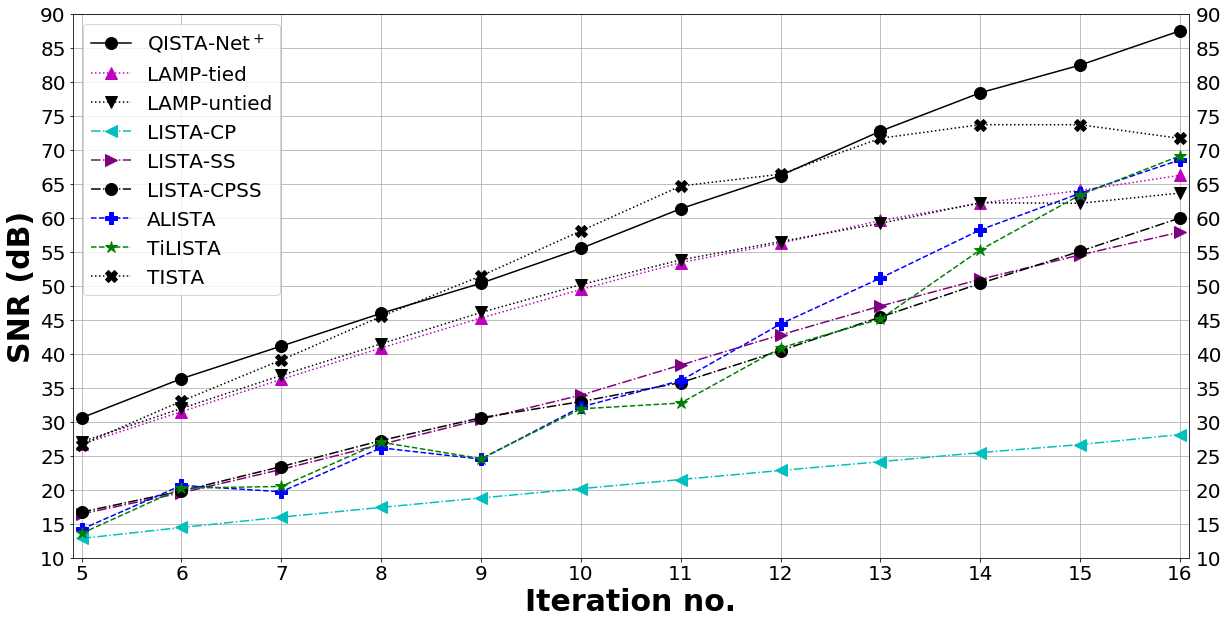}
\caption{Comparison in reconstructing the exactly $k$-sparse ground-truth,
with measurement rate 50\% ($n=500, m=250$).}
\label{compare_noiseless}
\end{center}
\end{figure}

\begin{figure}[!h]
\begin{center}
\includegraphics[width=\linewidth]{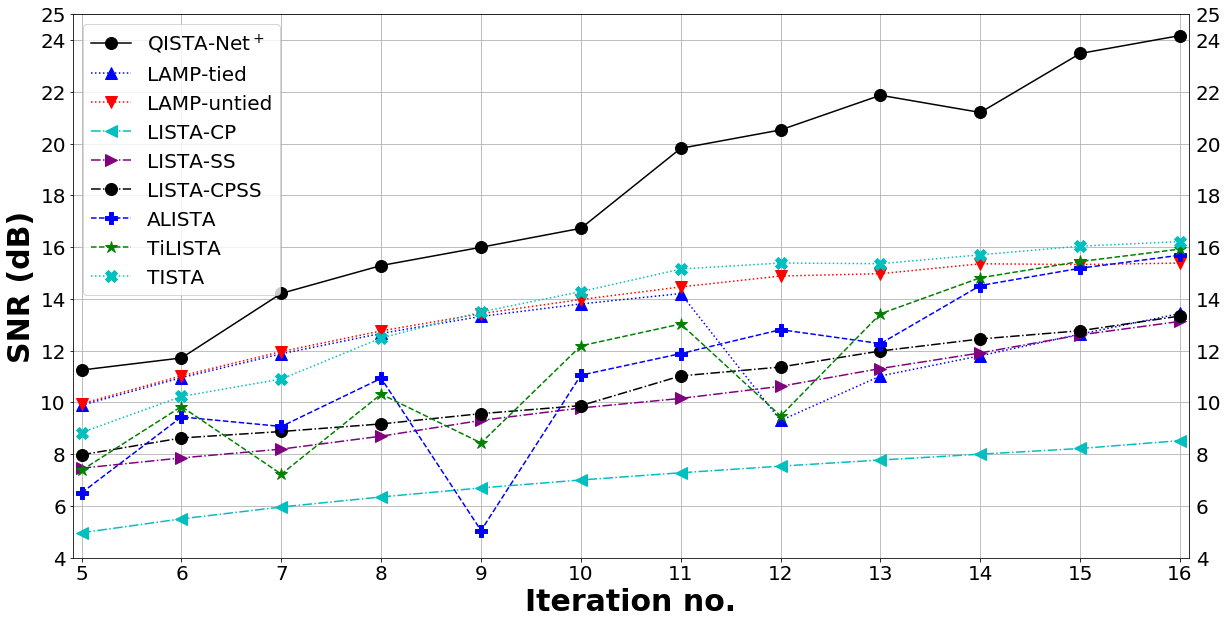}
\caption{Comparison in reconstructing the exactly $k$-sparse ground-truth,
with measurement rate 30\% ($n=500, m=150$).}
\label{compare_m_rate_150}
\end{center}
\end{figure}

\begin{figure}[!h]
\begin{center}
\includegraphics[width=\linewidth]{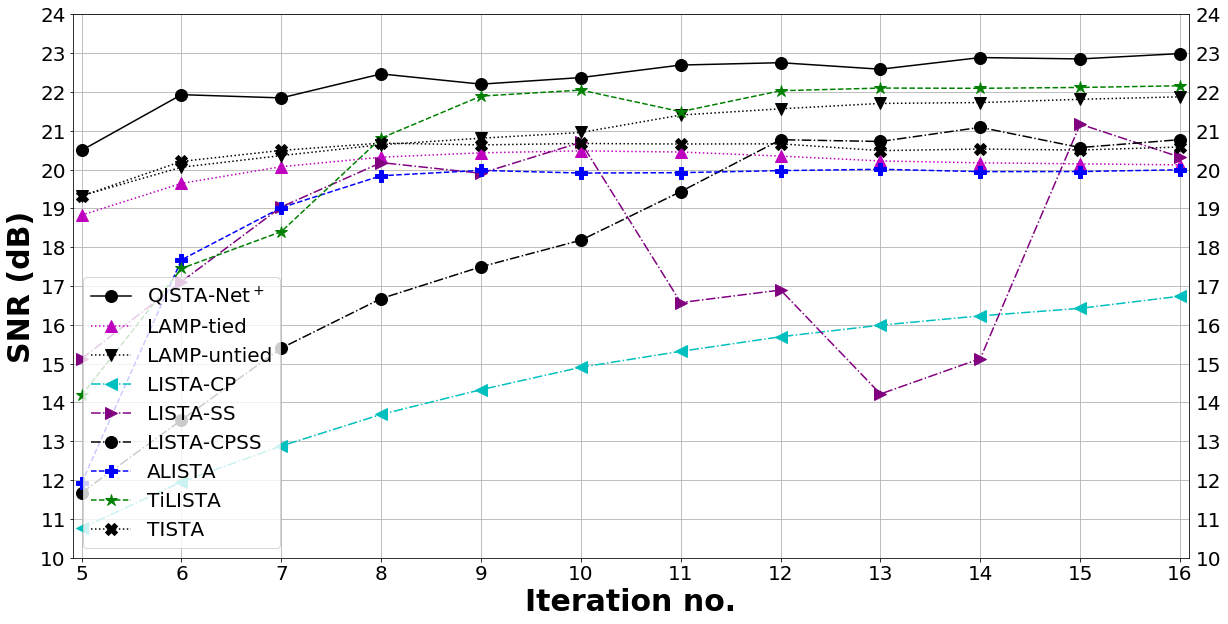}
\caption{Comparison in reconstructing the ground-truth
at measurement noise SNR=$20$dB with measurement rate 50\% ($n=500, m=250$).}
\label{compare_noisy_20dB}
\end{center}
\end{figure}

\begin{figure}[!hbpt]
\begin{center}
\includegraphics[width=\linewidth]{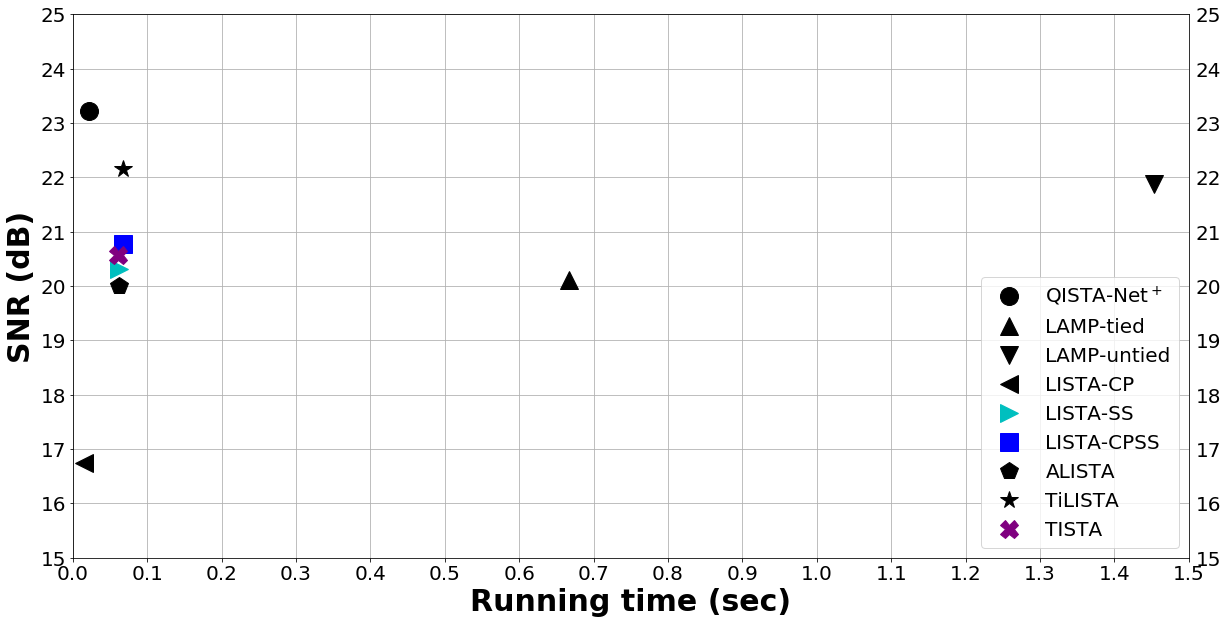} \caption{Running time vs. Reconstruction performance 
at measurement noise SNR=$20$dB with measurement rate 50\% ($n=500, m=250$) at 16$^{\mbox{\footnotesize{th}}}$ iteration/layer of Fig. \ref{compare_noisy_20dB}.}
\label{compare_testing_time}
\end{center}
\end{figure}

In this section, we conducted comparisons of QISTA-Net-s
with state-of-the-art deep learning-based sparse signal reconstruction
methods, including LAMP tied \cite{LVAMP} and LAMP untied \cite{LVAMP},
LISTA-CP \cite{LISTA-CPSS}, LISTA-SS \cite{LISTA-CPSS}, and LISTA-CPSS \cite{LISTA-CPSS},
ALISTA \cite{ALISTA} and TiLISTA \cite{ALISTA},
and TISTA \cite{TISTA}, in Fig. \ref{compare_noiseless}, Fig. \ref{compare_m_rate_150}, Fig. \ref{compare_noisy_20dB}, and Fig. \ref{compare_testing_time}.
Remark that we only compare with the $\ell_1$-based methods because, to our knowledge, we have not found any $\ell_q$-based ($0 < q < 1$) learning-based methods.
The results of LAMP-tied, LAMP-untied, LISTA-CP, LISTA-SS, LISTA-CPSS, TiLISTA, ALISTA, and TISTA were obtained by applying the codes provided by the corresponding authors.
Moreover, LISTA (which is the comparison target in \cite{LVAMP, LISTA-CPSS, ALISTA, TISTA}) was not selected because
all of the network models, we adopted to compare with, have already completely outperformed it
(including LISTA \cite{LISTA-LeCun} that shares the learning parameters $S$ and $B$ for each layer, and LISTA \cite{LVAMP} that learns $A^t$ and $B^t$ in each layer independently)
in terms of reconstruction quality.
The reconstruction quality was measured by Signal-to-Noise-Ratio (SNR $=10\log_{10}\left(\frac{\left\|x_0\right\|_2^2}{\left\|x^*-x_0\right\|_2^2}\right)$ dB).
All of the reconstruction results of QISTA-Net-s are of the averaged
at 100 tests.

In Fig. \ref{compare_noiseless}, we show the performance comparison between QISTA-Net-s and the other state-of-the-art $\ell_1$-based DNN methods, in reconstructing the exactly $k$-sparse ground-truth ($n=500, m=250$, and $k\approx50$).
We can see that QISTA in Sec. \ref{sec:QISTA} is the iterative method whereas QISTA-Net-s is its accelerated version via DNN architecture. At the
$16^{\mbox{\footnotesize th}}$ iteration (layer), the reconstruction performance of QISTA-Net-s is still good.
The performance comparison of reconstructing the exactly $k$-sparse ground-truth under measurement rate $30\%$ ($n=500, m=150$, and $k\approx50$) is shown in Fig. \ref{compare_m_rate_150}.
In Fig. \ref{compare_noisy_20dB}, we demonstrate the performance comparison in reconstructing the ground-truth with measurement noise at SNR=$20$dB ($n=500, m=250$, and $k\approx50$).
In general, the $\ell_q$-norm (and also $\ell_0$-norm) minimization problem is considered not robust, that is, if the original signal is noisy, the reconstruction result is considered to be unreliable (in comparison with the result of $\ell_1$-norm). But, as shown in Fig. \ref{compare_noisy_20dB}, QISTA-Net-s still outperforms all the $\ell_1$-norm learning-based methods, this is because the network architecture of QISTA-Net-s is also including the method to solve the $\ell_1$-norm problem.
Finally, in Fig. \ref{compare_testing_time}, 
we show the comparison results in terms of running time (sec.) vs. reconstruction performance (SNR (dB))
at measurement noise SNR=$20$dB under $n=500, m=250$, and $k\approx50$.

In summary, we can see from Fig. \ref{compare_noiseless}, Fig. \ref{compare_m_rate_150}, and Fig. \ref{compare_noisy_20dB} that QISTA-Net-s outperforms all the other existing works in
terms of reconstruction quality.
Moreover, in Fig. \ref{compare_m_rate_150}, all the $\ell_1$-based DNN methods used for comparison only achieve a maximum of $17$dB (at $16^{\mbox{\footnotesize{th}}}$ layer).
We conjecture that this is because, for the $\ell_1$-based iterative methods, $m$ must be greater than $3k$ to achieve a good reconstruction performance \cite{C07}.
Furthermore, Fig. \ref{compare_testing_time} actually indicates that QISTA-Net-s offers state-of-the-art performance in terms of reconstruction quality and speed.

\subsection{Ablation Study}\label{sec:Ablation Study QISTA-Net-s}

In the sparse signal reconstruction problem,
the $\left(\ell_0\right)$-problem (in Eq. (\ref{L0}))
is commonly used to find the sparse ground-truth.
Due to the NP-hard property of the $\left(\ell_0\right)$-problem,
it is always adopting LASSO (in Eq. (\ref{LASSO})) to
approximate the $\left(\ell_0\right)$-problem in the literature.
In this paper, we challenge the $\left(\ell_q\right)$-problem (in Eq. (\ref{Lq})), $0<q<1$,
which is non-convex, to approximate the $\left(\ell_0\right)$-problem.

\begin{figure}[!h]
\begin{center}
\includegraphics[width=\linewidth]{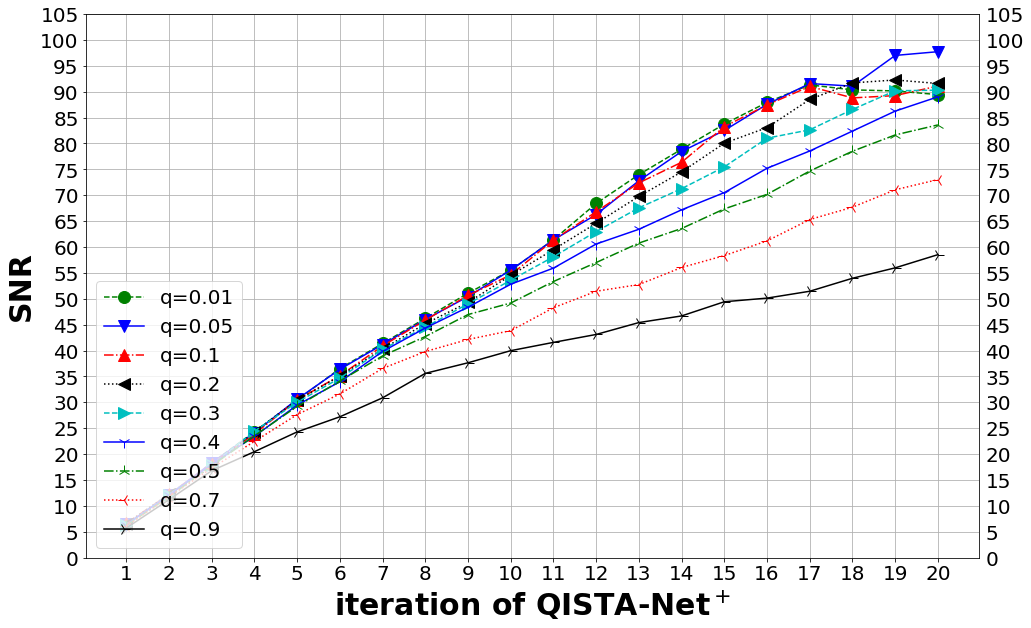}
\caption{QISTA-Net-s in reconstructing the exactly $k$-sparse ground-truth, with measurement rate 50\% ($n=500, m=250$), under various $q$'s.}
\label{QISTA-Nets-tune-q}
\end{center}
\end{figure}

In Sec. \ref{sec:QISTA-Net-s}, we designed a learning-based method,
QISTA-Net-s, to
solve the $\left(\ell_q\right)$-problem.
In Fig. \ref{QISTA-Nets-tune-q}, we demonstrate the reconstruction performance of QISTA-Net-s with respect to $q$.
The dimension setting is $n=500$ and $m=250$ (the measurement rate is 50\%).
The other parameter setting is followed in
Sec. \ref{sec:Parameter Setting QISTA-Net-s}.
We observe that reconstruction quality is increased when $q$ is decreased.
This indicates that in solving the $\left(\ell_q\right)$-problem is better approaches the
$\ell_0$-norm minimization problems than its $\ell_1$-norm counterpart, as suggested in \cite{C07, CY08}.

In Sec. \ref{sec:QISTA}, we proposed an iterative algorithm QISTA in solving the sparse signal reconstruction problem.
In Sec. \ref{sec:QISTA-Net-s}, we designed a DNN method by unfolding the parameters in QISTA, as shown in Eq. (\ref{eq:QISTA-unfold}), to accelerate the reconstruction process.
After unfolding QISTA, we further accelerated
by considering the momentum coming from the descent direction
of all previous layers, and the result was QISTA-Net-s (in Algorithm
\ref{alg:QISTA-Net-s}).

\begin{figure}[!h]
\begin{center}
\includegraphics[width=\linewidth]{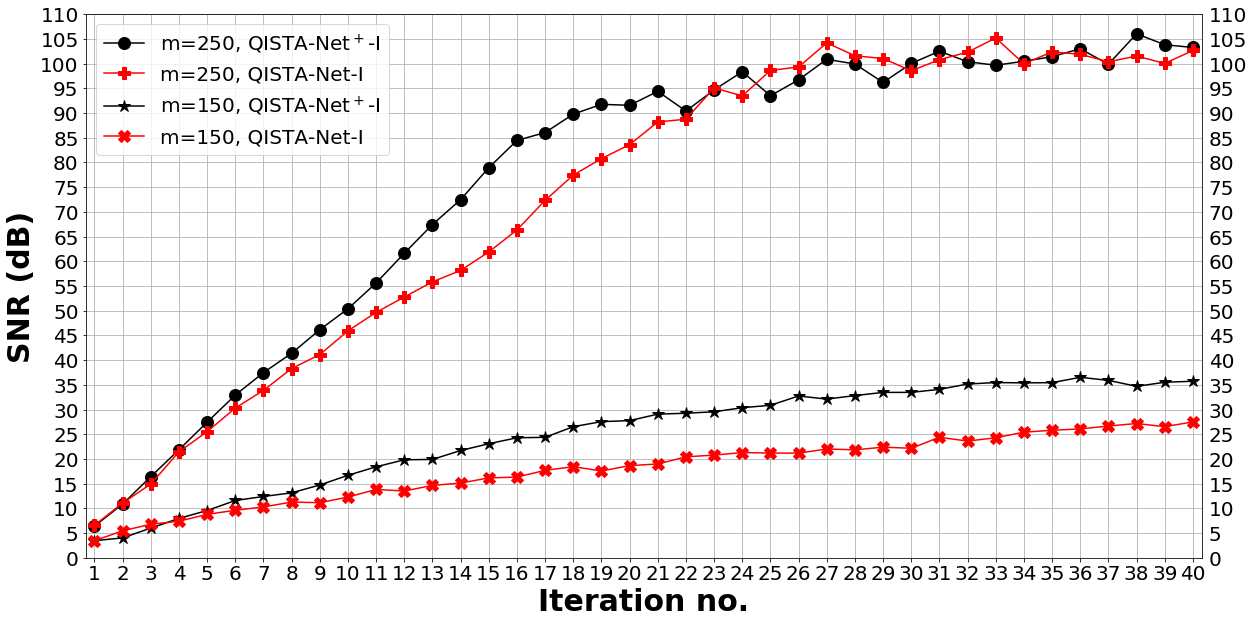}
\caption{Performance comparison between QISTA-Net-s and QISTA-unfold in reconstructing the exactly $k$-sparse ground-truth under measurement rates 50\% and 30\%.}
\label{fig:QISTA-Net-compare}
\end{center}
\end{figure}

In Fig. \ref{fig:QISTA-Net-compare},
we study the effect of the momentum coming from all
of the previous layers. That is, we compare the network
architecture in Algorithm \ref{alg:QISTA-Net-s}
with the network architecture in Eq. (\ref{eq:QISTA-unfold})
(we call QISTA-unfold for convenient).
The dimension setting is $n=500$. We compare the two tasks
in two measurement rates, $50\%$ ($m=250$), and  $30\%$ ($m=150$).
The other parameter setting is followed in
Sec. \ref{sec:Parameter Setting QISTA-Net-s}.
All of the results are presented in an average of $100$ tests.

As shown in Fig. \ref{fig:QISTA-Net-compare}, we can see that, under the same SNR values, QISTA-Net-s needs fewer layers
than QISTA-unfold.
As $m=250$, to reach SNR=$80$dB, QISTA-Net-s took only $16$ layers but QISTA-unfold needs $19$ layers,
and then the two have a similar reconstruction performance after 23$^{\mbox{\footnotesize rd}}$ layer.
Moreover, when $m=150$, the improvement of QISTA-Net-s is significant.
This indicates that the speed-up strategy of momentum is effective.

\section{Experimental Results of Deep Learning-Based
Method In Reconstructing the Natural Images}\label{sec:exp QISTA-Net-n}

In this section, we show the reconstruction performance of
QISTA-Net-n in Algorithm \ref{alg:QISTA-Net-n}
that reconstructing the natural images 
in the deep learning-based method and compare the performance
with state-of-the-art deep leaning-based methods.

\subsection{Parameter Setting}\label{sec:Parameter Setting QISTA-Net-n}

The constant parameter in  QISTA-Net-n was 
$\varepsilon=0.1\cdot\textbf{1}_n$.
The training parameters of QISTA-Net-n were initialized as
$\beta^t=0.1$, $\lambda^t=10^{-5}$, $\alpha^t=1$,
and the others $\left\{\mathcal{B}, \mathcal{A}, \mbox{ and } \mathcal{C}_i^t, i=0, 2, \cdots ,7\right\}$
were initialized by xavier initializer \cite{XAVIER}.

For the convolutional dictionaries $\Psi$ and $\tilde{\Psi}$, together with the extras $\mathcal{C}_0$ and $\mathcal{C}_7$, all the kernels of convolution operators were set to $3\times 3$,
and the numbers of input features and output features of
$\mathcal{C}_0, \mathcal{C}_1, \mathcal{C}_2, \mathcal{C}_3, \mathcal{C}_4,
\mathcal{C}_5, \mathcal{C}_6$, and $\mathcal{C}_7$ are listed in Table \ref{feature table}.

\begin{table}
\caption{The list of input feature and output feature.}\label{feature table}
\centering
\begin{tabular}[!h]{rcccccccc}
\hline
& $\mathcal{C}_0$ & $\mathcal{C}_1$ & $\mathcal{C}_2$ & $\mathcal{C}_3$ & $\mathcal{C}_4$ & $\mathcal{C}_5$ & $\mathcal{C}_6$ & $\mathcal{C}_7$\\
\hline
Input feature    &  1 & 32 & 32 & 32 & 32 & 32 & 32 & 32\\
Output feature   & 32 & 32 & 32 & 32 & 32 & 32 & 32 & 1
\end{tabular}
\end{table}

\subsection{Datasets for Training and Testing}
In the experiment to reconstruct the natural image, the training dataset is the Set91 (91 images) database; the Set11 (11 images) database is adopted for validation data; and the Set5 (5 images), Set14 (14 images), BSD68 (68 images), and BSD100 (100 images) databases are the testing datasets.

The training data is generated by cropping the $64\times64$ pixels sub-images with the stride of 7, of each image in Set91, into 68,323 sub-images. Moreover, the training data is augmented by flipped, rotation 90, rotation 90 plus flipped, rotation 180, rotation 180 plus flipped, rotation 270, and rotation 270 plus flipped, to each sub-image, with finally 546,584 sub-images.

\subsection{Training Details}

For a fair comparison, we adopt a similar training setting with SCSNet \cite{SCSNet}
(and also with CSNet$^+$ \cite{CSNet} because both of them has the same training setting)
that the measurement matrix $A$ operating on the image block with block size $32\times32$. This will require us to do the necessary pre-processing and post-processing
in step 2 of QISTA-Net-n.
We follow the training setting in SCSNet because it had the best reconstruction performance in the comparison object.
The training batch in the proceeding is in image-shape, in size of $64\times64\times b$,
where $b$ is the training batch size, and $64\times64$ is the size of a training data.
The required processing is as follows. To simplify the illustration, here we suppose the batch size $b=1$.
\begin{itemize}
\item[1.]
In step 2 of QISTA-Net-n, since the operator $\mathcal{A}$
is unfolding by $A\in\mathbb{R}^{m\times1024}$,
and the input $x^{t-1}$ of step 2 has the size of $64\times 64$,
we first divide the sub-image $x^{t-1}$ into $4$ blocks, each of such has the size of $32\times 32$.
Next, we reshape each of the blocks into vector-form $\mathbb{R}^{1024}$ (4 vectors: $x_1, x_2, x_3, x_4$),
then multiple each of $x_i$ by $\mathcal{A}\in\mathbb{R}^{m\times 1024}$ (the measurement rate $\frac{m}{1024}$),
and get the results $y_i=\mathcal{A}x_i\in\mathbb{R}^{m}$, $i=1,2,3,4$.
The size of the object in the process is as shown in the following.
\begin{eqnarray}\label{eq:A_process}
{\small\hspace*{-15pt}\begin{array}{c@{\ \ }c@{\ \ }c@{\ \ }c@{\ \ }c@{\ \ }c@{\ \ }c}
64\times64
&\xrightarrow{\mbox{\footnotesize divide}}& 32\times32\times 4
&\rightarrow&1024\times 4
&\xrightarrow{\mathcal{A}}& m\times4
\end{array}}
\end{eqnarray}

\item[2.]
For operator $\mathcal{B}$ in step 2, we see that
the size of $y-\mathcal{A}x^{t-1}$ is $m\times4$.
Since $\mathcal{B}$ is unfolding by $A^T\in\mathbb{R}^{1024\times m}$,
we have $\mathcal{B}\left(y-\mathcal{A}x^{t-1}\right)\in\mathbb{R}^{1024\times4}$.
After the operating of $\mathcal{B}$, we reverse the pre-processing that we did before $\mathcal{A}$, which leads the results back into the image-form of the size $64\times64$.
The size of the object in the process we do is as shown in the following.
\begin{eqnarray}\label{eq:B_process}
{\small\hspace*{-15pt}\begin{array}{c@{\ \ }c@{\ \ }c@{\ \ }c@{\ \ }c@{\ \ }c@{\ \ }c}
m\times 4&\xrightarrow{\mathcal{B}}& 1024\times4
&\rightarrow&32\times32\times 4
&\xrightarrow{\mbox{\footnotesize merge}}& 64\times64
\end{array}}
\end{eqnarray}
\end{itemize}

The pre-process as in Eq. (\ref{eq:A_process}), and the post-process as in Eq. (\ref{eq:B_process}) is to reduce the memory cost of measurement matrix $\mathcal{A}$
(we will adopt $\mathcal{A}$ as the measurement matrix after the training finished).
When we sense an image, assuming that the measurement rate is $r_0$, no matter whether the image block size is $64\times64$ or $32\times32$, the sampling rate is the same, but the required storage cost for measurement matrix $\mathcal{A}$ is different.
If the image block size is $64\times64$, $\mathcal{A}$ has the size of $m\times4096$,
whereas if the image block size is $32\times32$, $\mathcal{A}$ has the size of $m\times1024$.

The training data consists of 273,292 sub-images in size of $64\times64$ pixels.
In each batch, we randomly choose 64 sub-images from the training data to train.
Remark that the image block size we adopt in this paper is $64\times64$
instead of $96\times96$ as in SCSNet \cite{SCSNet} due to the restriction of device memory.

\subsection{Performance Comparison of Reconstructing The Natural Image}\label{sec:Performance Comparison of Natural Image}

The experiments in this subsection were conducted on
Intel Core i7-7700 CPU plus a NVIDIA GeForce GTX 1080 Ti GPU, Python with TensorFlow version 1.14.0.

In this subsection, we compare the reconstruction performance with
state-of-the-art learning-based methods 
including SDA \cite{SDA}, ReconNet \cite{ReconNet},
ISTA-Net$^+$ \cite{ISTA-Net},
MS-CSNet \cite{MS-CSNet}, DR$^2$-Net \cite{DR2-Net},
$\{0,1\}$-BCSNet \cite{CSNet}, $\{-1,+1\}$-BCSNet \cite{CSNet},
CSNet$^+$ \cite{CSNet}, SCSNet \cite{SCSNet}, MSRNet \cite{MSRNet},
method in \cite{DBCS}, method in \cite{IRCNN}, and method in \cite{KC2019}.
The comparison results are shown in Table \ref{comparison-Set11},
Table \ref{comparison-BSD68}, Table \ref{comparison-Set5},
Table \ref{comparison-Set14}, and Table \ref{comparison-BSD100},
which are corresponding to the datasets Set11, BSD68, Set5, Set14, and BSD100, respectively.
In each dataset, the best reconstruction results are marked in bold font.
The reconstruction results of the comparison objects (in PSNR and SSIM)
are obtained by the experiment results of the corresponding papers.
The ``dash'' mark in the tables means the authors did not provide.

Table \ref{comparison-Set11} shows the comparison of reconstruction performance
in terms of PSNR of different learning-based methods on various measurement rates.
Because Set11 is adopted as the validation data, QISTA-Net-n obtains better
reconstruction performance in terms of PSNR than the
comparison objects therein.

In Table \ref{comparison-BSD68}, we compare the reconstruction performance
in terms of PSNR with other learning-based methods.
In Table \ref{comparison-Set5}, 
Table \ref{comparison-Set14}, and Table \ref{comparison-BSD100},
we compare the reconstruction performance in both PSNR and SSIM
with other learning-based methods. 
The reconstruction performance of QISTA-Net-n in terms of PSNR outperforms
the other methods in all measurement rates therein, except the
$1\%$ measurement rate.
For $1\%$ measurement rate, SCSNet has the best reconstruction performance in PSNR. 
We guess this is because SCSNet adopts the sub-images in size of $96\times96$ pixels as the input of the network architecture,
and this implies that SCSNet has less blocking artifact effect.
However, QISTA-Net-n cannot take the sub-images in the size of $96\times96$ pixels as the training data due to limitations on the memory size of the GPUs.
On the other hand, the reconstruction performance of QISTA-Net-n in terms of SSIM, although not better than all of the other methods, but it has almost the same as the best one.

\begin{figure*}
\centering
\begin{subfigure}{0.25\textwidth}
\centering
\includegraphics[width=1.0\linewidth,keepaspectratio]{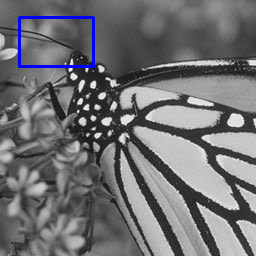}
\caption{``Monarch'' in dataset Set11}
\end{subfigure}
\begin{subfigure}{0.18\textwidth}
\centering
\includegraphics[width=1.0\linewidth]{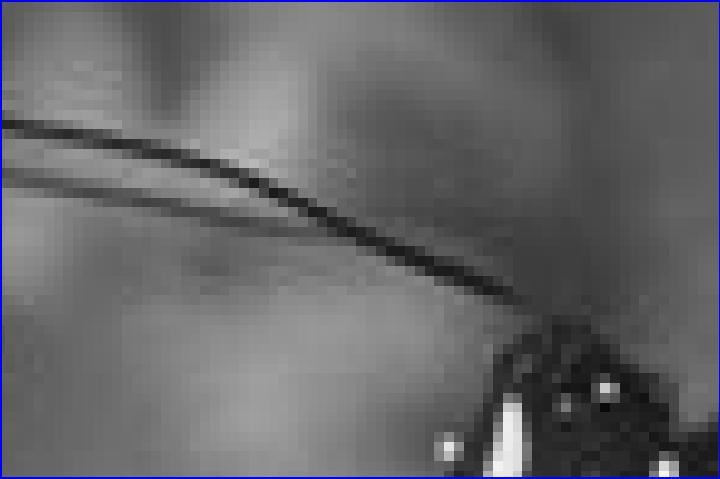}
\caption{ground-truth}
\end{subfigure}
\begin{subfigure}{0.18\textwidth}
\centering
\includegraphics[width=1.0\linewidth]{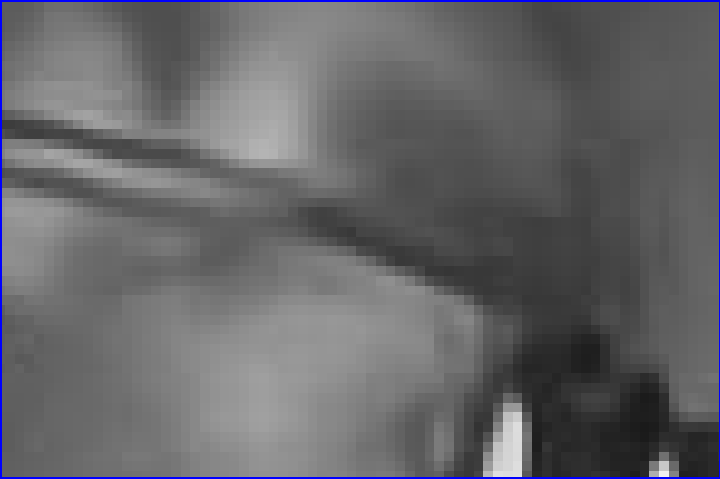}
\caption{CSNet$^+$}
\end{subfigure}
\begin{subfigure}{0.18\textwidth}
\centering
\includegraphics[width=1.0\linewidth]{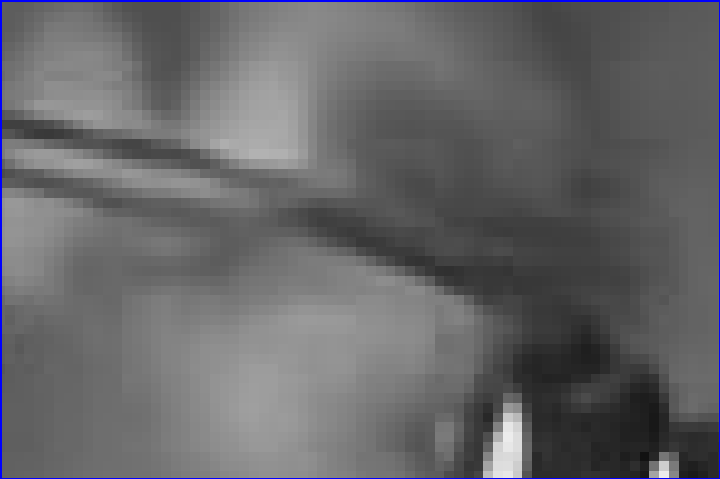}
\caption{SCSNet}
\end{subfigure}
\begin{subfigure}{0.18\textwidth}
\centering
\includegraphics[width=1.0\linewidth]{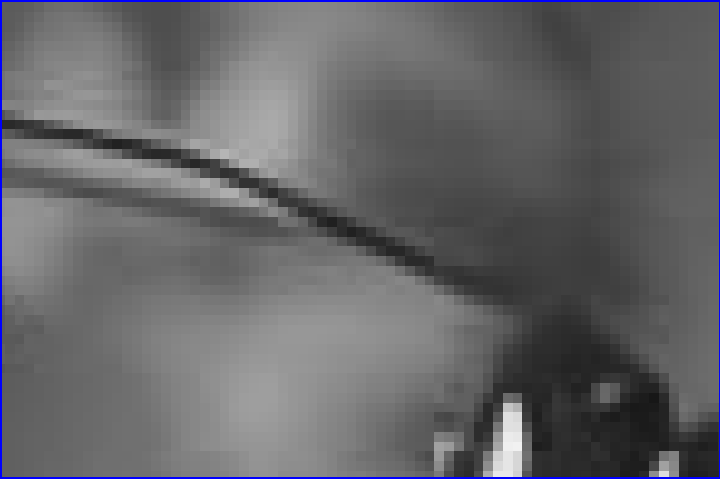}
\caption{QISTA-Net-n}
\end{subfigure}
\caption{Reconstruction Result of CSNet$^+$, SCSNet, and QISTA-Net-n with 10\% measurement rate.}
\label{fig:Monarch}
\end{figure*}

\begin{figure*}
\centering
\begin{subfigure}{0.25\textwidth}
\centering
\includegraphics[width=1.0\linewidth,keepaspectratio]{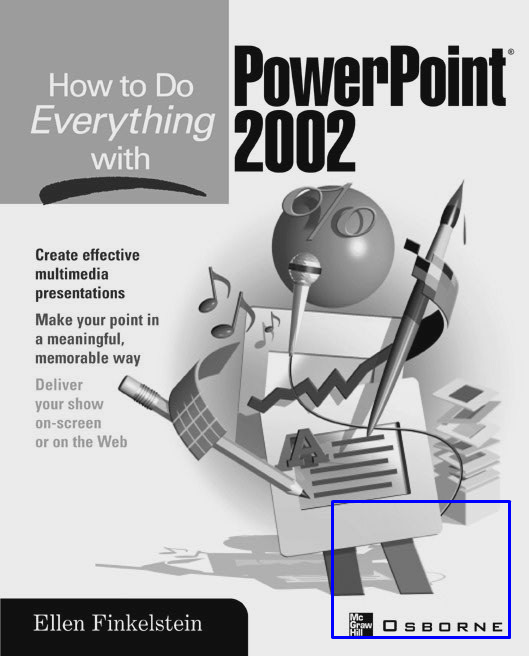}
\caption{``ppt3'' in dataset Set14}
\end{subfigure}
\begin{subfigure}{0.18\textwidth}
\centering
\includegraphics[width=1.0\linewidth]{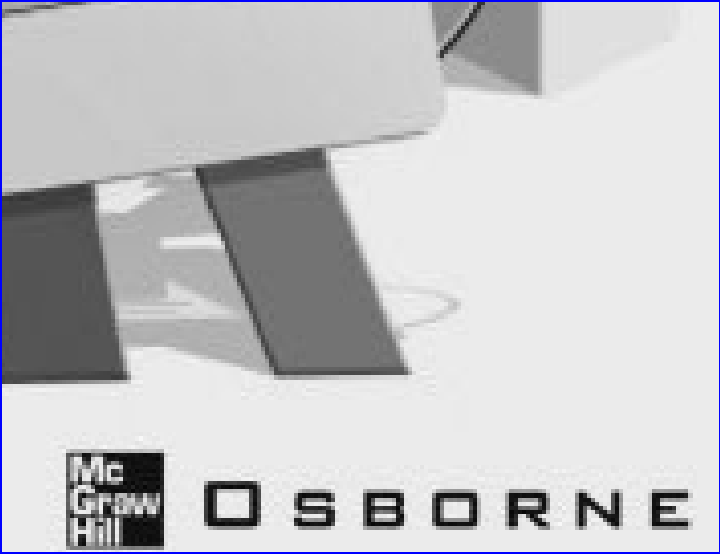}
\caption{ground-truth}
\end{subfigure}
\begin{subfigure}{0.18\textwidth}
\centering
\includegraphics[width=1.0\linewidth]{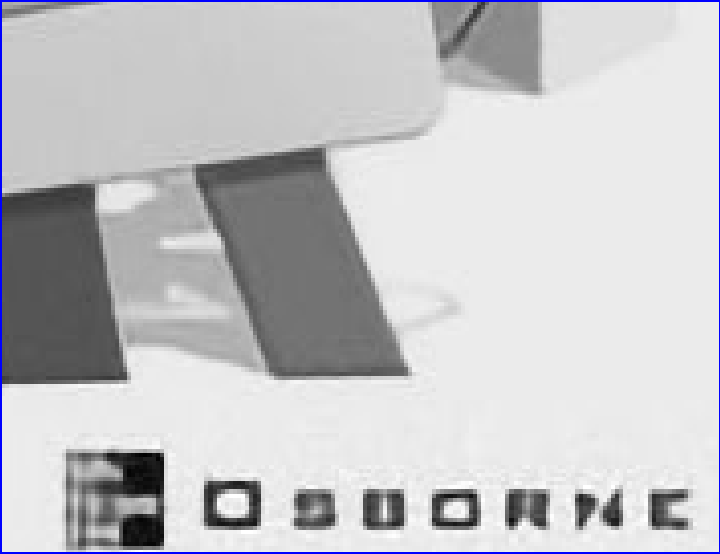}
\caption{CSNet$^+$}
\end{subfigure}
\begin{subfigure}{0.18\textwidth}
\centering
\includegraphics[width=1.0\linewidth]{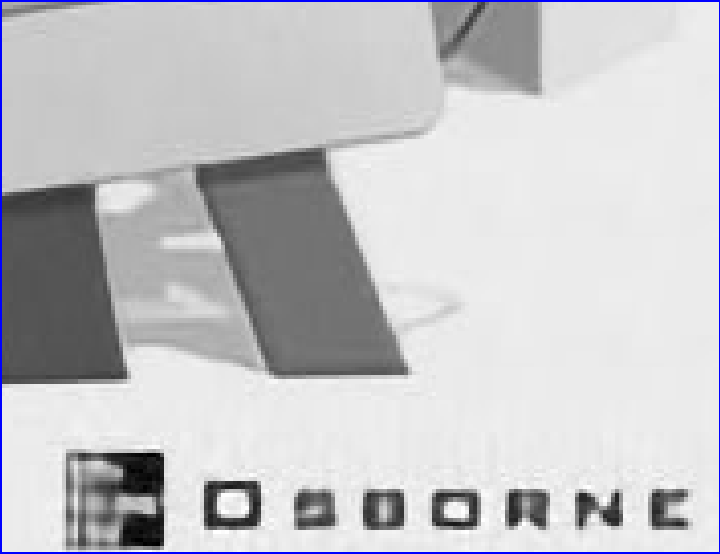}
\caption{SCSNet}
\end{subfigure}
\begin{subfigure}{0.18\textwidth}
\centering
\includegraphics[width=1.0\linewidth]{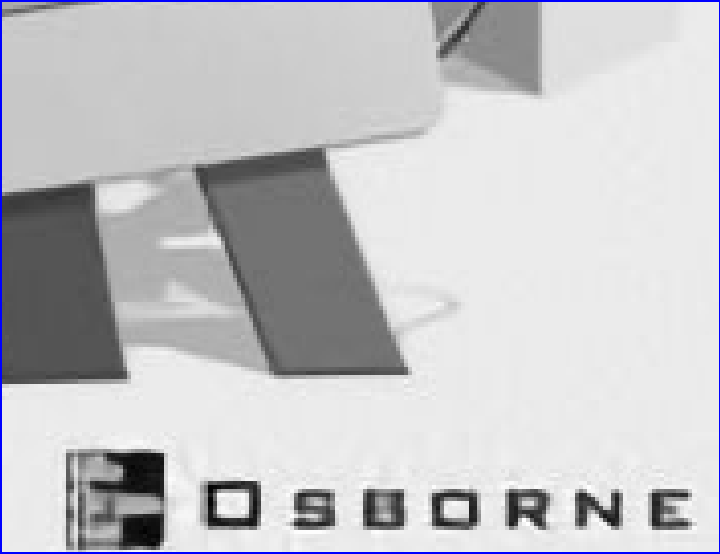}
\caption{QISTA-Net-n}
\end{subfigure}
\caption{Reconstruction Result of CSNet$^+$, SCSNet, and QISTA-Net-n with 10\% measurement rate.}
\label{fig:ppt3}
\end{figure*}

\begin{figure*}
\centering
\begin{subfigure}{0.25\textwidth}
\centering
\includegraphics[width=1.0\linewidth,keepaspectratio]{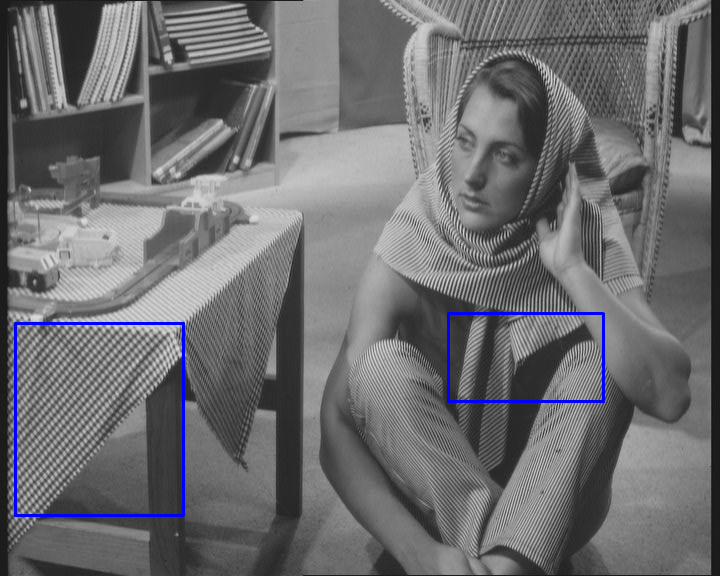}
\caption{``babara'' in dataset Set14}
\end{subfigure}
\begin{subfigure}{0.18\textwidth}
\centering
\includegraphics[width=1.0\linewidth]{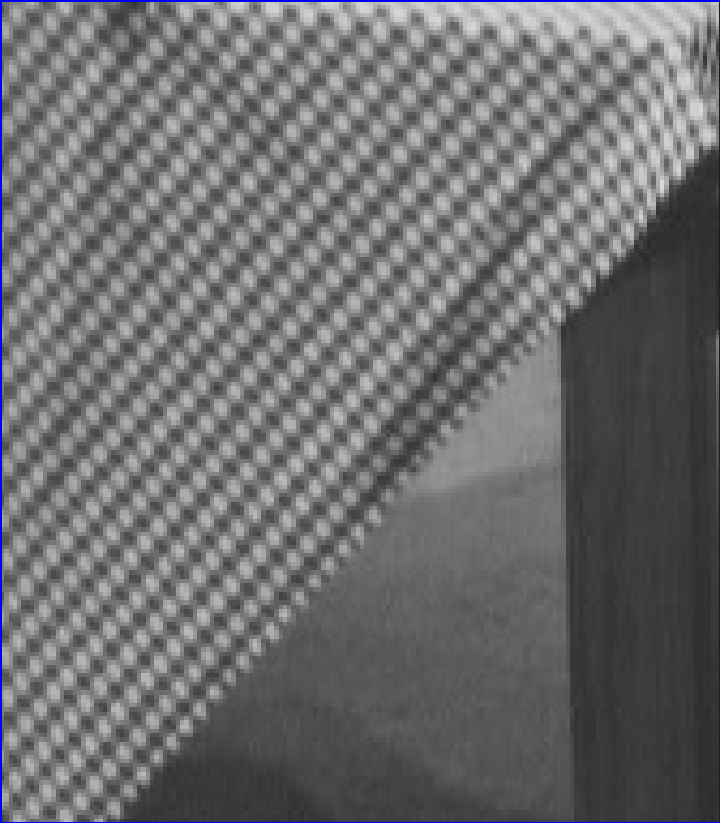}
\caption{ground-truth}
\end{subfigure}
\begin{subfigure}{0.18\textwidth}
\centering
\includegraphics[width=1.0\linewidth]{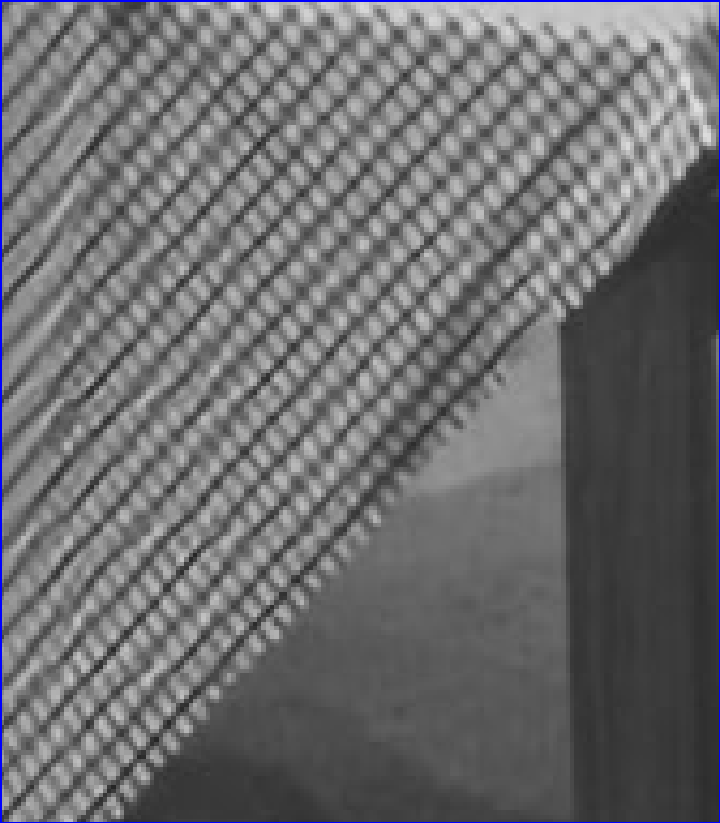}
\caption{CSNet$^+$}
\end{subfigure}
\begin{subfigure}{0.18\textwidth}
\centering
\includegraphics[width=1.0\linewidth]{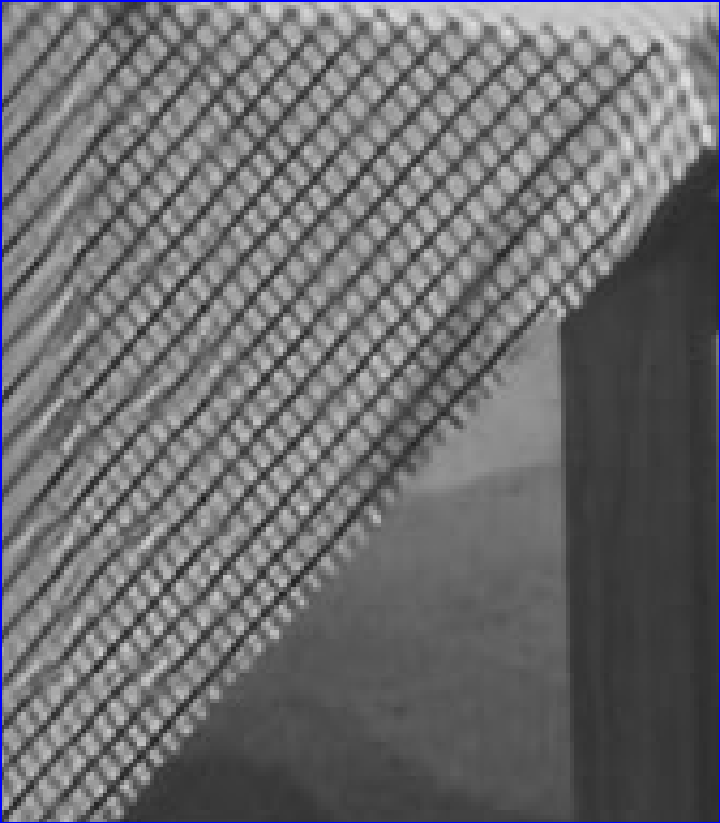}
\caption{SCSNet}
\end{subfigure}
\begin{subfigure}{0.18\textwidth}
\centering
\includegraphics[width=1.0\linewidth]{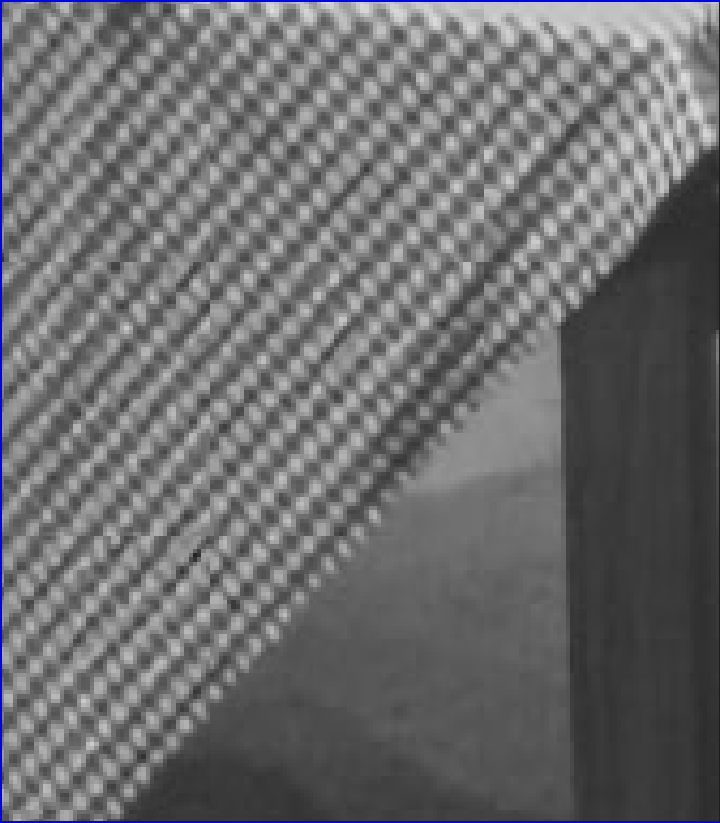}
\caption{QISTA-Net-n}
\end{subfigure}

\hspace*{+126pt}
\begin{subfigure}{0.18\textwidth}
\centering
\includegraphics[width=1.0\linewidth]{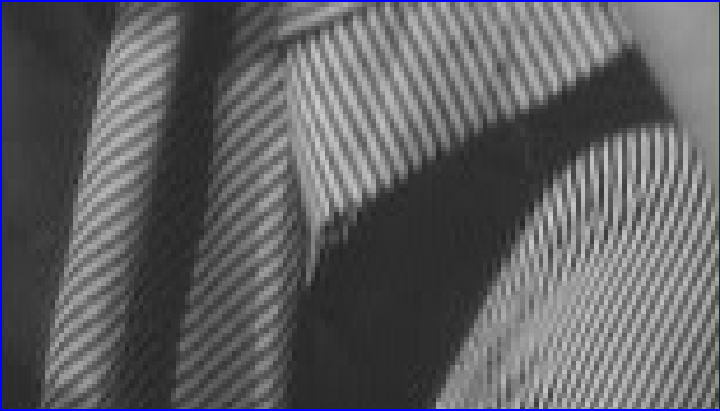}
\caption{ground-truth}
\end{subfigure}
\begin{subfigure}{0.18\textwidth}
\centering
\includegraphics[width=1.0\linewidth]{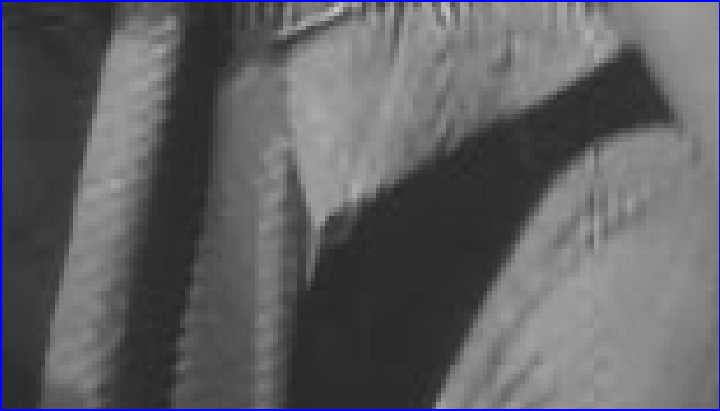}
\caption{CSNet$^+$}
\end{subfigure}
\begin{subfigure}{0.18\textwidth}
\centering
\includegraphics[width=1.0\linewidth]{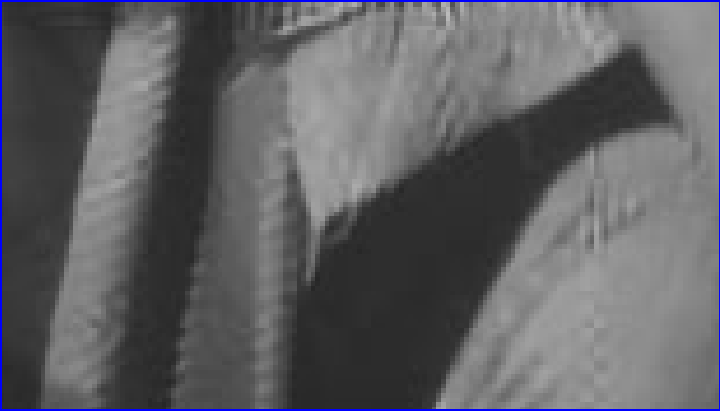}
\caption{SCSNet}
\end{subfigure}
\begin{subfigure}{0.18\textwidth}
\centering
\includegraphics[width=1.0\linewidth]{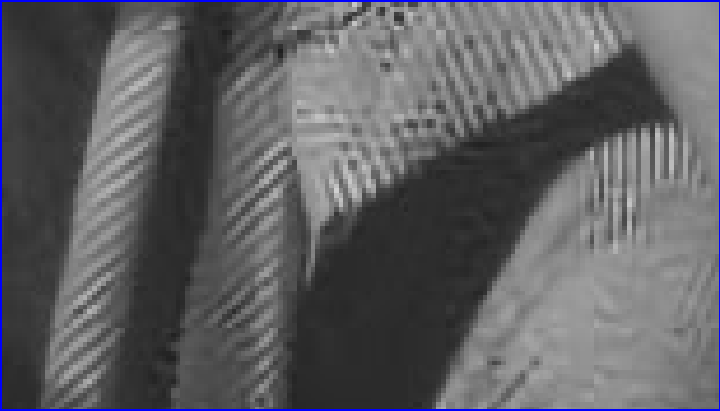}
\caption{QISTA-Net-n}
\end{subfigure}
\caption{Reconstruction Result of CSNet$^+$, SCSNet, and QISTA-Net-n with 20\% measurement rate.}
\label{fig:barbara}
\end{figure*}

Fig. \ref{fig:barbara}, Fig. \ref{fig:Monarch}, and Fig. \ref{fig:ppt3} show the visual
comparison between ground-truth and the reconstruction results of CSNet$^+$, SCSNet, and
QISTA-Net-n.
We do not compare with the other methods because either the authors do not provide the implementation codes or the visual comparison in \cite{SCSNet} is already done with obviously comparison results.
Besides, we compare with only CSNet$^+$ and SCSNet because the comparison results of these two methods almost overwhelming the other related works.
``Monarch'' in Fig. \ref{fig:Monarch} is from the dataset Set11, whereas ``ppt3'' in Fig. \ref{fig:ppt3} and ``barbara'' in Fig. \ref{fig:barbara} are from the dataset Set14.
Fig. \ref{fig:Monarch} show the comparison results between CSNet$^+$, SCSNet, and
QISTA-Net-n, in reconstruct the ``Monarch'' image from the dataset Set14, with measurement rate 10\%.
We can see that the reconstruction result of QISTA-Net-n at the Butterfly's tentacles is relatively not blurred.
Fig. \ref{fig:ppt3} show the results in reconstruct the ``ppt3'' image from the dataset Set14, with measurement rate 10\%.
We can clearly recognize the letter E at the reconstruction result of QISTA-Net-n.
Fig. \ref{fig:barbara} show the results in reconstruct the ``barbara'' image from the dataset Set14, with measurement rate 20\%.
In the reconstruction result of QISTA-Net-n, the texture of the tablecloth is closer to the original pattern.
On the contrary, the other two methods are a bit sharp in the upper right-lower left texture.
Moreover, in the comparison between Fig. \ref{fig:barbara} (g)-(i), we can see that QISTA-Net-n tries to reconstruct the striped texture.


\begin{table*}
\caption{Average PSNR (dB) comparisons of different methods with
various measurement rates on Set11.}\label{comparison-Set11}
\centering
\begin{tabular}[t]{c|ccccccc}
\hline
measurement rate & $50\%$ & $40\%$ & $30\%$ & $25\%$ & $10\%$ & $4\%$ & $1\%$\\
\hline
SDA \cite{SDA}&28.95&27.79&26.63&25.34&22.65&20.12&17.29\\
ReconNet \cite{ReconNet}&31.50&30.58&28.74&25.60&24.28&20.63&17.27\\
\cite{IRCNN} &36.23&34.06&31.18&30.07&24.02&17.56&7.70\\
LISTA-CPSS \cite{LISTA-CPSS}&34.60&32.87&30.54&-&-&-&-\\
ISTA-Net$^+$ \cite{ISTA-Net}&38.07&36.06&33.82&32.57&26.64&21.31&17.34\\
DR$^2$-Net \cite{DR2-Net}&-&-&-&29.06&24.71&21.29&17.80\\
$\{0,1\}$-BCSNet \cite{CSNet}&35.05&34.61&32.57&-&26.39&-&20.62\\
$\{-1,+1\}$-BCSNet \cite{CSNet}&35.57&34.94&33.42&-&28.03&-&20.93\\
CSNet$^+$ \cite{CSNet}&38.52&36.48&34.30&-&28.37&-&21.03\\
SCSNet \cite{SCSNet}&39.01&36.92&34.62&-&28.48&-&21.04\\
MSRNet \cite{MSRNet}&-&-&-&33.36&28.07&24.23&20.08\\
\cite{KC2019}&-&-&-&32.81&26.97&-&18.83\\
\cdashline{1-8}[0.8pt/2pt]
QISTA-Net-n &\textbf{40.30}&\textbf{38.32}&\textbf{36.09}&\textbf{34.84}&\textbf{29.79}&\textbf{25.95}&\textbf{21.29}\\
\hline
\end{tabular}
\end{table*}

\begin{table*}
\caption{Average PSNR (dB) comparisons of different methods with
various measurement rates on BSD68.}\label{comparison-BSD68}
\centering
\begin{tabular}[t]{c|ccccccc}
\hline
measurement rate & $50\%$ & $40\%$ & $30\%$ & $25\%$ & $10\%$ & $4\%$ & $1\%$\\
\hline
SDA \cite{SDA}&28.35&27.41&26.38&-&23.12&21.32&-\\
ReconNet \cite{ReconNet}&29.86&29.08&27.53&-&24.15&21.66&-\\
ISTA-Net$^+$ \cite{ISTA-Net}&34.01&32.21&30.34&-&25.33&22.17&-\\
CSNet \cite{CSNet}&34.89&32.53&31.45&-&27.1&-&22.34\\
SCSNet \cite{SCSNet}&35.77&33.86&31.87&-&27.28&-&\textbf{22.37}\\
\cite{KC2019}&-&-&-&29.54&-&-&-\\
\cdashline{1-8}[0.8pt/2pt]
QISTA-Net-n &\textbf{36.56}&\textbf{34.58}&\textbf{32.64}&\textbf{31.63}&\textbf{27.83}&\textbf{25.24}&22.23\\
\hline
\end{tabular}
\end{table*}

\begin{table*}
\caption{Average PSNR (dB) and SSIM comparisons of different methods with
various measurement rates on Set5.}\label{comparison-Set5}
\hspace*{-20pt}
\begin{tabular}[t]{c|c@{\ \ }c|c@{\ \ }c|c@{\ \ }c|c@{\ \ }c|c@{\ \ }c|c@{\ \ }c|c@{\ \ }c}
\hline
Set5 dataset&PSNR&SSIM&PSNR&SSIM&PSNR&SSIM&PSNR&SSIM
&PSNR&SSIM&PSNR&SSIM&PSNR&SSIM\\
\hline
measurement rate & \multicolumn{2}{c|}{$50\%$} & \multicolumn{2}{c|}{$40\%$} & \multicolumn{2}{c|}{$30\%$} & \multicolumn{2}{c|}{$20\%$} & \multicolumn{2}{c|}{$10\%$} & \multicolumn{2}{c|}{$5\%$} & \multicolumn{2}{c}{$1\%$}\\
\hline
ReconNet \cite{ReconNet}&-&-&-&-&-&-&-&-&25.98&0.734&-&-&-&-\\
\cite{DBCS} &-&-&-&-&36.54&0.956&34.55&0.939&31.31&0.894&-&-&-&-\\
MS-CSNet \cite{MS-CSNet}&-&-&-&-&38.43&\textbf{0.966}&36.26&0.950&32.82&0.909&-&-&-&-\\
DR$^2$-Net \cite{DR2-Net}&-&-&-&-&-&-&-&-&27.79&0.798&-&-&-&-\\
$\{0,1\}$-BCSNet \cite{CSNet}&38.69&0.970&38.24&0.967&36.44&0.955&32.31&0.898&
29.99&0.851&28.57&0.816&23.79&0.636\\
$\{-1,+1\}$-BCSNet \cite{CSNet}&39.23&0.967&38.62&0.964&37.22&0.956&35.24&0.939&
32.20&0.898&29.39&0.840&24.07&0.645\\
CSNet$^+$ \cite{CSNet}&41.79&0.980&40.11&0.974&38.25&0.964&36.05&0.948&32.59&0.906&
29.74&0.849&24.18&0.648\\
SCSNet \cite{SCSNet}&42.22&\textbf{0.982}&40.44&\textbf{0.976}&38.45&\textbf{0.966}&36.15&0.949&32.77&0.908&
29.74&0.847&\textbf{24.21}&0.647\\
\cdashline{1-15}[0.8pt/2pt]
QISTA-Net-n &\textbf{43.17}&\textbf{0.982}&\textbf{41.27}&0.975&\textbf{39.24}&\textbf{0.966}&\textbf{37.04}&\textbf{0.951}&\textbf{33.63}&\textbf{0.917}&\textbf{30.23}&\textbf{0.863}&23.60&\textbf{0.650}\\
\hline
\end{tabular}
\end{table*}

\begin{table*}
\caption{Average PSNR (dB) and SSIM comparisons of different methods with
various measurement rates on Set14.}\label{comparison-Set14}
\hspace*{-20pt}
\begin{tabular}[t]{c|c@{\ \ }c|c@{\ \ }c|c@{\ \ }c|c@{\ \ }c|c@{\ \ }c|c@{\ \ }c|c@{\ \ }c}
\hline
Set14 dataset&PSNR&SSIM&PSNR&SSIM&PSNR&SSIM&PSNR&SSIM
&PSNR&SSIM&PSNR&SSIM&PSNR&SSIM\\
\hline
measurement rate & \multicolumn{2}{c|}{$50\%$} & \multicolumn{2}{c|}{$40\%$} & \multicolumn{2}{c|}{$30\%$} & \multicolumn{2}{c|}{$20\%$} & \multicolumn{2}{c|}{$10\%$} & \multicolumn{2}{c|}{$5\%$} & \multicolumn{2}{c}{$1\%$}\\
\hline
ReconNet \cite{ReconNet}&-&-&-&-&-&-&-&-&24.18&0.640&-&-&-&-\\
\cite{DBCS} &-&-&-&-&33.08&0.926&31.21&0.885&28.54&0.814&-&-&-&-\\
MS-CSNet \cite{MS-CSNet}&-&-&-&-&34.34&0.930&32.26&0.896&29.29&0.820&-&-&-&-\\
DR$^2$-Net \cite{DR2-Net}&-&-&-&-&-&-&-&-&24.38&0.706&-&-&-&-\\
$\{0,1\}$-BCSNet \cite{CSNet}&35.01&0.945&34.52&0.938&32.68&0.912&29.25&0.816&
27.36&0.756&26.09&0.694&22.48&0.553\\
$\{-1,+1\}$-BCSNet \cite{CSNet}&35.56&0.944&34.81&0.934&33.47&0.916&31.55&0.880&
28.78&0.805&26.67&0.724&22.74&0.562\\
CSNet$^+$ \cite{CSNet}&37.89&0.963&36.16&0.950&34.34&0.930&32.15&0.894&29.13&0.817&
26.93&0.733&22.83&0.563\\
SCSNet \cite{SCSNet}&38.41&0.966&36.54&0.953&34.51&0.931&32.19&0.895&29.22&0.818&
26.92&0.732&\textbf{22.87}&0.563\\
\cdashline{1-15}[0.8pt/2pt]
QISTA-Net-n &\textbf{39.61}&\textbf{0.969}&\textbf{37.57}&\textbf{0.955}&\textbf{35.49}&\textbf{0.936}&\textbf{33.18}&\textbf{0.903}&\textbf{29.97}&\textbf{0.829}&\textbf{27.51}&\textbf{0.747}&22.63&\textbf{0.564}\\
\hline
\end{tabular}
\end{table*}

\begin{table*}
\caption{Average PSNR (dB) and SSIM comparisons of different methods with
various measurement rates on BSD100.}\label{comparison-BSD100}
\hspace*{-20pt}
\begin{tabular}[t]{c|c@{\ \ }c|c@{\ \ }c|c@{\ \ }c|c@{\ \ }c|c@{\ \ }c|c@{\ \ }c|c@{\ \ }c}
\hline
BSD100 dataset&PSNR&SSIM&PSNR&SSIM&PSNR&SSIM&PSNR&SSIM
&PSNR&SSIM&PSNR&SSIM&PSNR&SSIM\\
\hline
measurement rate & \multicolumn{2}{c|}{$50\%$} & \multicolumn{2}{c|}{$40\%$} & \multicolumn{2}{c|}{$30\%$} & \multicolumn{2}{c|}{$20\%$} & \multicolumn{2}{c|}{$10\%$} & \multicolumn{2}{c|}{$5\%$} & \multicolumn{2}{c}{$1\%$}\\
\hline
MS-CSNet \cite{MS-CSNet}&-&-&-&-&33.19&\textbf{0.919}&31.15&0.874&28.61&0.786&-&-&-&-\\
$\{0,1\}$-BCSNet \cite{CSNet}&33.95&0.937&33.41&0.928&31.67&0.894&28.65&0.785&
27.05&0.722&26.04&0.658&23.49&0.541\\
$\{-1,+1\}$-BCSNet \cite{CSNet}&34.57&0.940&33.67&0.925&32.28&0.899&30.50&0.855&
28.21&0.770&26.55&0.689&23.70&0.547\\
CSNet$^+$ \cite{CSNet}&36.68&0.962&34.91&0.944&33.08&0.917&31.05&0.872&28.53&0.783&
26.78&0.698&23.76&\textbf{0.548}\\
SCSNet \cite{SCSNet}&37.14&\textbf{0.965}&35.21&\textbf{0.947}&33.24&\textbf{0.919}&31.10&0.873&28.57&0.784&
26.77&0.697&\textbf{23.78}&\textbf{0.548}\\
\cdashline{1-15}[0.8pt/2pt]
QISTA-Net-n &\textbf{37.55}&0.964&\textbf{35.60}&0.946&\textbf{33.66}&\textbf{0.919}&\textbf{31.53}&\textbf{0.875}&\textbf{28.87}&\textbf{0.788}&\textbf{26.95}&\textbf{0.703}&23.46&0.543\\
\hline
\end{tabular}
\end{table*}

\subsection{Running Time Comparison of Reconstructing The Natural Images}

Table \ref{running-time} show the average running time in reconstructing a $256\times256$ image.
The running time of the comparison objects are taken from \cite{CSNet} and \cite{SCSNet}.
We can see that the running time of QISTA-Net-n in GPU is close to CSNet$^+$, but the performance is better than CSNet$^+$ in most cases.
Besides, in comparison with SCSNet, Qista-Net-n also obtains better performance in most cases, but the running time of Qista-Net-n is 5 times faster than CS.

\begin{table*}
\caption{Average running time (in seconds) of different methods for reconstructing
a $256\times256$ image.}\label{running-time}
\begin{center}
\begin{tabular}[t]{c|cc|cc|c}
\hline
measurement rate & \multicolumn{2}{c|}{$1\%$} & \multicolumn{2}{c|}{$10\%$}
& \multirow{2}{*}{Programming Language}\\
\cdashline{1-5}[0.8pt/2pt]
& CPU & GPU & CPU & GPU&\\
\hline
SDA \cite{SDA} & - & 0.0045 & - & 0.0029 & \multirow{2}{*}{Matlab + caffe}\\
ReconNet \cite{ReconNet} & 0.5193 & 0.0244 & 0.5258 & 0.0195 &\\
\cdashline{1-6}[0.8pt/2pt]
ISTA-Net$^+$ \cite{ISTA-Net} & 1.3750 & 0.0470 & 1.3750 & 0.0470 & Python + TensorFlow\\ 
\cdashline{1-6}[0.8pt/2pt]
$\{0,1\}$-BCSNet \cite{CSNet} & 0.9066 & 0.0275 & 0.9081 & 0.0281
& \multirow{3}{*}{Matlab + Matconvnet}\\
$\{-1,+1\}$-BCSNet \cite{CSNet} & 0.9115 & 0.0274 & 0.9065 & 0.0276 &\\
CSNet$^+$ \cite{CSNet} & 0.8960 & 0.0262 & 0.9024 & 0.0257 &\\
\cdashline{1-6}[0.8pt/2pt]
SCSNet \cite{SCSNet} & 0.5103 & 0.1050 & 0.5146 & 0.1332 & Matlab + Matconvnet \\
\cdashline{1-6}[0.8pt/2pt]
QISTA-Net-n &0.7620&0.0255&0.7575&0.0238& Python + TensorFlow \\
\hline
\end{tabular}
\end{center}
\end{table*}

\section{Conclusion and Future Work}\label{sec:conclusion}

In this paper, we first reformulate the $\ell_q$-norm minimization problem
into a $2$-step problem (\ref{2-step}),
which excludes the difficulties coming from the non-convexity.

We then propose QISTA to solve the $\left(\ell_q\right)$-problem via the $\ell_1$-norm based iterative algorithm.
The QISTA obtains reasonable result associated with suggestion of $\left(\ell_q\right)$-problem in that the smaller $q$ leads to the better reconstruction performance.
Moreover, with the help of deep learning strategy and ``momentum'' strategy,
we propose an $\ell_q$ learning-based method, QISTA-Net-s.
The resultant QISTA-Net-s retains better reconstruction performance and faster reconstruction speed than state-of-the-art $\ell_1$-norm DNN methods, even if the original sparse signal is noisy.
Finally, with the help of convolution neural network, we proposed the QISTA-Net-n to solve the image CS problem.
The reconstruction performance outperforms most of the state-of-the-art
natural images reconstruction methods.


\bibliographystyle{IEEEbib}
\bibliography{refs}

\end{document}